\pdfoutput=1
\documentclass[11pt]{article} 

\usepackage[final]{acl} 

\usepackage{times} 
\usepackage{latexsym} 
\usepackage{enumitem} 
\usepackage{lipsum} % 用于生成示例文本

\usepackage[T1]{fontenc} 
\usepackage[utf8]{inputenc} 
\usepackage{array} 
\usepackage{booktabs} 
\usepackage{microtype} 
\usepackage{multirow} 
\usepackage{inconsolata} 

\usepackage{amsmath} 
\title{Deciphering the Impact of Pretraining Data on Large Language Models through Machine Unlearning} 

\author{Yang Zhao$^{\spadesuit}\footnotemark[2]$, Li Du $^{\heartsuit}\footnotemark[2]
  $,$\,$ Xiao Ding$^{\spadesuit}\footnotemark[1]$, $\,$ Kai Xiong$^{\spadesuit}$, $\,$ Zhouhao Sun$^{\spadesuit}$, Jun Shi$^{\clubsuit}$, Ting Liu $^{\spadesuit}$and Bing Qin$^{\spadesuit}$ \\
  $^{\spadesuit}$Research Center for Social Computing and Information Retrieval\\ %
  Harbin Institute of Technology, China \\
  $^{\heartsuit}$Beijing Academy of Artificial Intelligence, Beijing, China\\
  $^{\clubsuit}$Academy of Cyber, Beijing, China\\
 \tt{\{yangzhao, xding, kxiong, hzsun, tliu, qinb\}@ir.hit.edu.cn}\\
 \tt{duli@baai.ac.cn} \,\, \
 \tt{junshi1770@gmail.com}\\,\\}
\usepackage{graphicx} 
\begin{document} 
\maketitle
\renewcommand*{\thefootnote}{\fnsymbol{footnote}}
\footnotetext[2]{These authors contributed equally to this work.}
\renewcommand*{\thefootnote}
{\fnsymbol{footnote}}
\footnotetext[1]{Corresponding Author.}
\renewcommand*{\thefootnote}
{\arabic{footnote}}
\begin{abstract} 

Through pretraining on a corpus with various sources, Large Language Models (LLMs) have gained impressive performance. However, the impact of each component of the pretraining corpus remains opaque. As a result, the organization of the pretraining corpus is still empirical and may deviate from the optimal. To address this issue, we systematically analyze the impact of 48 datasets from 5 major categories of pretraining data of LLMs and measure their impacts on LLMs using benchmarks about nine major categories of model capabilities. 
Our analyses provide empirical results about the contribution of multiple corpora on the performances of LLMs, along with their joint impact patterns, including complementary, orthogonal, and correlational relationships. We also identify a set of  ``high-impact data'' such as Books that is significantly related to a set of model capabilities. These findings provide insights into the organization of data to support more efficient pretraining of LLMs.

% Our analyses unveil the unseen effects of algorithms on the Textual Entailment, Mathematical Reasoning task, along with the complementary, orthogonal, and correlational relationships among datasets, as well as a set of "high-impact data" that. These findings provide insights into the organization of data to support more efficient pretraining of Large Language Models (LLMs).

% Our analyses unveil the 未见的作用of xxx xxx xxx on yyy yyy yyy任务, 以及数据集之间的互补、正交、相关关系, as well as a set of "high-impact data" that zzz。Providing insights about the organization of to support more efficient pretraining of LLMs.

\end{abstract}

\section{Introduction} 

Under the data-driven paradigm, Large Language Models (LLMs) have demonstrated promising performance and showcased immense potential in further promotion  \cite{openai2023gpt4, touvron2023llama2, du2021glm, bai2023qwen}. 
%Through the pre-training process on corpora with vast scale, diverse source (e.g., Github, Books, Arxiv, etc.), and various types (e.g., text, math, or code), LLMs can acquire abundant knowledge and powerful capability in understanding the natural language, wielding commonsense knowledge, and solving complex reasoning tasks, such as math and coding problems. 
Previous analyses have suggested that the composition of the pretraining corpus may exert a significant impact upon the performance of LLMs \cite{longpre2023pretrainer, shen2023slimpajama}. 
However, how different sources and types of pertaining corpora influence the \emph{knowledge and reasoning ability} of LLMs largely remains opaque, or stays at the qualitative level. As a result, it still heavily relies on the experiences of trainers to organize the pre-training corpus. Such experiences may deviate from the optimal, and hence limit the efficiency and effectiveness of model training. 

%as different sources of data generally contain different content, and may differently contribute to the knowledge and ability of LLMs

%The core strength of these models lies in their self-supervised pre-training on extensive datasets, enhancing their ability to comprehend complex language and perform advanced reasoning tasks \cite{kaplan2020scaling, hoffmann2022empirical}. These models excel in tasks across various domains, including mathematical problem-solving \cite{wei2022chain} and code generation \cite{li2022competition}. However, despite previous analyses highlighting the impact of domain-specific content on model performance \cite{longpre2023pretrainer, shen2023slimpajama}, it remains challenging to explain the relationship between large model pre-training corpora and performance \cite{wang2023data}. Consequently, the organization of pre-training corpora still heavily relies on trainers' subjective experience, potentially far from optimal. 

In this paper, we propose to quantify how components with different sources and types in the pretraining corpus contribute to the performance of LLMs. 
Previous literature refers to such analyses as Data Influence Analysis (DIA) \cite{akyurek2022towards}. 
However, due to the limitations of previous DIA methods, the DIA of LLMs remains challenging. Primary DIA methods can be mainly categorized into two lines: the retraining-based methods and gradient-based methods. Retraining-based methods work by removing specific data from the training corpus and retraining the model, then comparing changes in model performance. Considering the prohibitive training cost of LLMs, retraining-based methods would be impractical \cite{nguyen2023bayesian}. While the gradient-based methods may not be applicable in analyzing the source of the complex reasoning ability of LLMs, as they assume that the performance upon a test instance is determined by several independent training instances.
%, and find such instances by comparing the gradient similarities between training and test instances \cite{koh2017understanding, hara2019data, hammoudeh2022training}. 
However, such an assumption may not hold for LLMs, especially for the ability to complete reasoning tasks, as it may originate from groups of correlated instances that jointly contribute to the performance of LLMs. For example, solving math problems requires understanding a knowledge taxonomy, and the taxonomy is described by a set of interdependent instances holistically. Missing one component would lead to the collapse of the whole taxonomy. Hence, the gradient-based methods may fail to trace the influence of such a whole corpus, which is of vital importance \cite{grosse2023studying}. 

%知识与**能力**的来源

% This study introduces a novel, structured approach to investigate how the content of pre-training corpora influences the performance of LLMs in downstream tasks.This study introduces a novel, structured approach to investigate how the content of pre-training corpora influences the performance of LLMs in downstream tasks..In previous research work, this task is referred to as training data attribution (TDA) \cite{akyurek2022towards}. 
% Two principal categories have been identified in TDA methodologies \cite{hammoudeh2022training}. 
% The first, Retraining-Based approaches, assess influence by training models with and without specific data instances \cite{aha1991instance, feldman2020neural, ghorbani2019data}. These approaches have primarily been applied to traditional machine learning models \cite{han2020explaining, hara2019data} and deep learning models with fewer parameters \cite{nguyen2023bayesian, akyurek2022towards}. The second category, Gradient-Based methods, involve comparing gradient similarities between training and test instances \cite{koh2017understanding, hara2019data, hammoudeh2022training}. Although analyses focusing on common sense memory or memorization samples provide relatively precise results, they are less effective in explaining tasks that require a comprehensive knowledge system, such as mathematical problem-solving and code development. These tasks often involve complex problem-solving and combinatorial generalization, where traditional analysis methods may prove insufficient \cite{grosse2023studying}.

\begin{figure} [t]
\centering
\setlength{\belowcaptionskip}{-10pt}
\includegraphics[width=\linewidth]{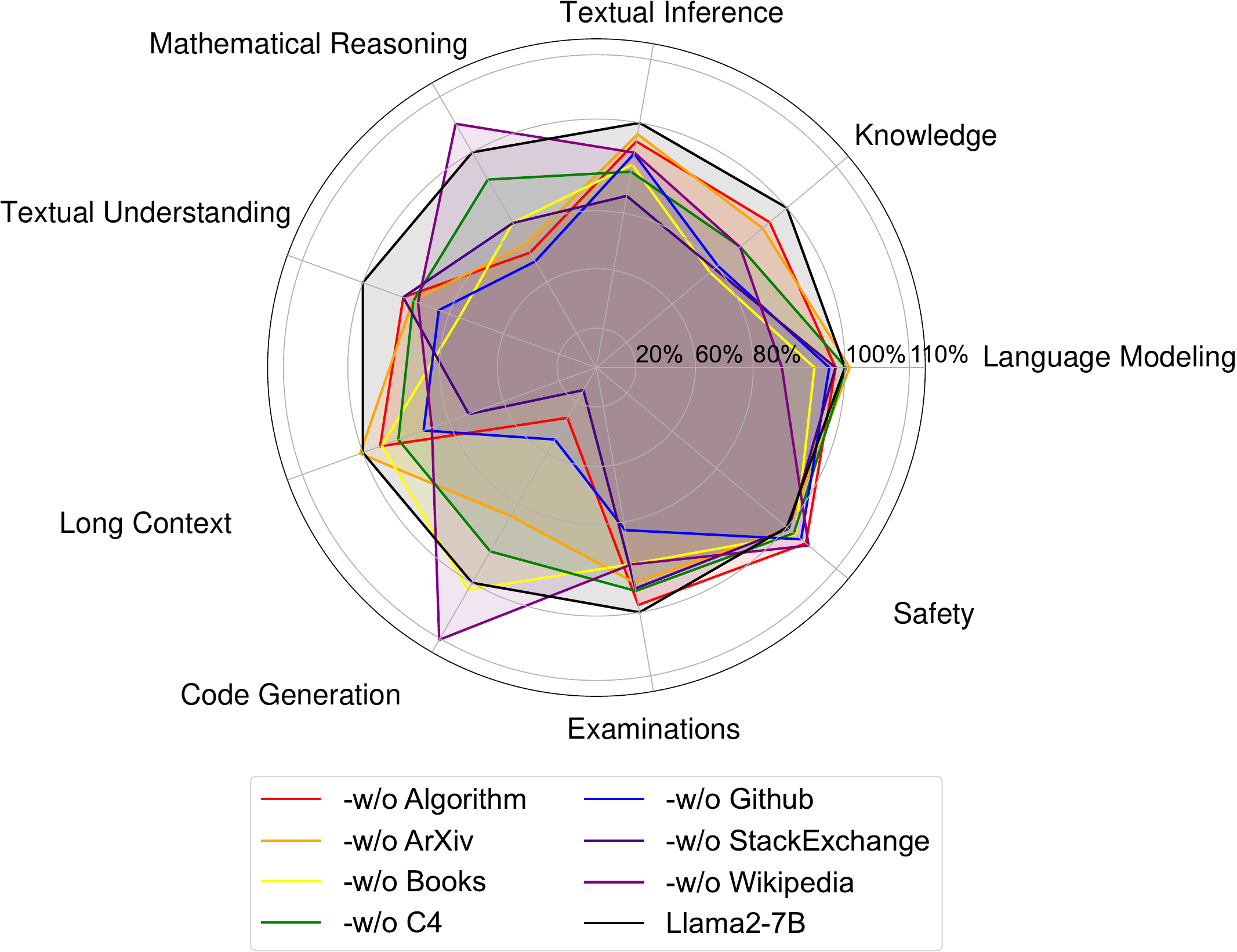} 
\caption{The impact of Unlearning different types of corpus on different abilities of the llama2-7B model.} 
\label{Figure 1} 
\end{figure}

To address these issues, we resort to another strand of method, Machine Unlearning. Prior research \cite{eldan2023s, jang2022knowledge} suggests that Machine Unlearning can selectively erase specific knowledge from a model through gradient \emph{ascent} on corresponding instances. This enables us to investigate the influence of a certain pretraining corpus on an LLMs by ``Unlearning'' instances from it, and then compare the performance of the ``forgotten'' LLMs with the original LLMs. Meanwhile, different from previous Machine Unlearning methods, to avoid unintended impacts on non-targeted samples, we incorporate additional regularization by retraining samples from non-targeted domains during the Unlearning process. 
%This methodology allows for assessing the influence of specific sample sets on the model by comparing the performance of the forgotten model with the original model on test instances.
Experiments demonstrate that our method can effectively remove information contained in the target samples, without significantly affecting other unrelated samples.

% 我们首先描述方法, 其次实验验证指向性。
% 原则：untargeted PPL不变, 遗忘领域：梯度上升是训练过程的逆转。

Based on our customized Machine Unlearning method, we systematically investigated the quantitative contribution of multiple important resources and types of training corpora on the performance of LLMs. 
% As listed in Table~\ref{table:Target_Data}, 
We covered widely adopted high-quality corpora GitHub, Wikipedia, ArXiv, Books, StackExchange and C4, and conducted an in-depth analysis of their contributions to the model performance by segmenting them into subsets based on the type of knowledge. From the content dimension, the abovementioned corpora covered text, commonsense knowledge, domain-specific knowledge, math, and coding. Additionally, to investigate the source of the reasoning abilities of LLMs, we analyzed the impact of over ten kinds of programming languages with different coding paradigms, and 17 kinds of common algorithms such as Dynamic Programming.
Programming paradigms essentially represent different abstractions of real-world problems, and an algorithm corresponds to a common solution for a particular type of reasoning problems. To comprehensively assess the impact of these corpora, we evaluated the ``forgotten'' LLMs across various downstream tasks as illustrated in Figure ~\ref{Figure 1} and detailed illustrated in Figure~\ref{result}.

% The primary contributions of our paper are as follows:
% %\begin{itemize} [label=\textbullet,itemsep=2pt, parsep=1pt, leftmargin=*, itemindent=1em]
% We propose an effective \textbf{GR} adient \textbf{A} s\textbf{C} ent-based Machine Unlearning method.
With our proposed gradient ascent-based Machine Unlearning method, we identify a range of corpora that have a significant impact on the capabilities of LLMs, and reveal previously unreported influences of certain corpus (such as algorithms) on model capabilities.
Furthermore, we discovered the interaction among corpora which jointly affects the capabilities of LLMs, and the existence of ``high-energy'' corpora, like Books in the data.  Our research underscores the importance of further studying the impact of pre-training data, to provide foundations for future research on the optimization of pre-training datasets to support a more efficient pre-training process.

Based on our analysis results, we provide several hints about the organization of the pretraining corpus, such how should the proportion of each category of
pretraining corpus be set, the arrangement of datasets during
the pretraining process, and the evaluation of the pretraining process of the LLMs.
%% Experimental results suggest that:

% \begin{itemize} [leftmargin=*, align=left]
% \begin{itemize} [leftmargin=*, align=left]
% \noindent\textbullet\ \ \begin{minipage} [t]{\dimexpr\textwidth-\parindent} 
% \setlength{\parindent} {\parindent} % 保持原有的段落缩进
% \setlength{\hangindent} {\parindent} % 设置悬挂缩进的大小
% \hangafter=1 % 从第一行开始应用悬挂缩进 
% \begin{itemize} [label=\textbullet,itemsep=2pt, parsep=1pt, leftmargin=*, itemindent=1em]

% \item
% An investigation into the distinct roles of various types of knowledge on model performance was conducted. It was found that, compared to programming languages, algorithms play a crucial role in affecting the model's mathematical reasoning and code generation capabilities.

% \item  High-impact data affecting multiple datasets were identified, including BOOK, the C programming language, ``highly-liked'' StackExchange entries, Shell, etc. Additionally, the presence of complementary, antagonistic, and orthogonal data within the pre-training corpus was discovered to significantly influence the training of the model.

% \item  It was observed that tasks of a comprehensive nature demand a wide variety of data types and do not rely on specific datasets. Conversely, tasks involving complex mathematical reasoning and code generation require a smaller range of data types, yet are significantly influenced by certain datasets.
% % \end{itemize} 
%     % \item 这是一个带黑点的项目, 首行缩进。
%     % \item 另一个带黑点的项目, 同样首行缩进。
% \end{itemize} 
\section{Methodology and Validity Analysis} 
\subsection{Machine Unlearning Based Data Influence Analysis} 
% Inspired by \citet{eldan2023s, jang2022knowledge}, 我们采用了基于梯度上升的machine Unlearning method for eliminating xxx. Moreover, to ensure the precision of the xxx, we further introduce a xxx regularization, and devise a (名字).
Following \citet{eldan2023s} and \citet{jang2022knowledge}, we devise our approach based on the Machine Unlearning to eliminate certain kinds of information from LLMs.. Formally, given an LLMs $M$ and a sample set $\mathcal{D}_T$, we analyze the influence of $\mathcal{D}_T$ upon $M$ by making $M$ ``unlearn'' the information of $\mathcal{D} _T$ to derive a model $M^u_T$, and then compare the performance between $M$ and $M^u_T$. For clarity, we call $M^u_T$ as the forgotten model, $\mathcal{D} _T$ as the targeted corpus, and the rest parts of the whole pretraining corpora as the non-targeted corpora.

The key of Machine Unlearning-based data influence analysis lies in that: (1) Effectiveness: How to unlearn $\mathcal{D}_T$ to make that $M$ has never been trained upon $\mathcal{D} _T$. (2) Precision: Do not incur unintentional impacts upon the non-targeted parts of the training corpus. To this end, we devise an \textbf{GR}adient \textbf{A}s\textbf{C}ent-based Machine Unlearning with r\textbf{E}\text{-}training (GRACE). In the following sections, we describe the mechanism of GRACE and show the effectiveness and precision of GRACE.

% Machine Unlearning Data  Ablation Influence Analysis (MUDAIA).
\subsection{Gradient-based Machine Unlearning with Retraining } 

During the training process, LLMs learn knowledge by \textbf{maximize} the likelihood of training corpora through gradient descent. Hence, in line with \citet{eldan2023s, jang2022knowledge}, the information within a targeted corpus $\mathcal{D} _T$ could be unlearned by reverting the learning process through gradient \emph{ascent} on $\mathcal{D} _T$. Formally, the objective function of the Unlearning algorithm is to \textbf{minimize} the log-likelihood upon $\mathcal{D} _T$.
% \begin{equation} \
% \small
%   Obj (M, \mathcal{D} _T)=-\sum_{i\in \mathcal{D} _T} \log (p_{M} (x))
% \end{equation} 
% where $x$ is an instance within $\mathcal{D} _T$, and $p_{M} (x)$ is the probability of $x$ modeled by $M$. 

%\subsubsection{Retrain Algorithm} 
% In this study, to mitigate any unintended impacts on performance related to the non-target corpus during the Unlearning process, we have introduced an additional regularization strategy. Formally, given a token sequence \ ( x = (x_1, \ldots, x_T) \), the training objective of our retrain algorithm is minimizing the log-likelihood:
However, there remains the risk that the performances on non-target domains are unintentionally impacted.
To avoid this problem, GRACE introduces an additional retraining regularization. 
Specifically, the information within a non-targeted corpus $\mathcal{D}_N$ could be revised through gradient descent upon $\mathcal{D}_N$. 
% Formally, the objective function of the retraining algorithm is to \textbf{minimize} the log-likelihood upon the target corpus:
% \begin{equation} \
% \small
%   Obj (M, \mathcal{D} _N)=-\sum_{i\in \mathcal{D} _N} \log (p_{M} (x))
% \end{equation} 
% where $x$ is an instance within $\mathcal{D} _N$, and $p_{M} (x)$ is the probability of $x$ modeled by $M$.
% \begin{equation} \
%   L_{UL} (f_\theta, x)=\textbf{-} \sum_{t=1} ^{T} \log (p_\theta (x_t|x<t))
% \end{equation} 
% where \ ( p_\theta (x_t|x<t) \) denotes the conditional probability of predicting the next token to be \ ( x_t \) when given \ ( x<t \) to an LM \ ( f \) with parameters \ ( \theta \). 

Hence, the whole algorithm runs in the following manner. Before starting, we first divide the  \emph{non-target} corpus $\mathcal{D}_N$ into a 9:1 split as a \emph{retraining set} and a \emph{dev set}. Then during the Unlearning process, if the model $M$'s Perplexity (PPL) on the \emph{dev set} is higher than that before Unlearning, a retraining is started until the PPL of $M$ on the \emph{dev set} restore to the original level on the \emph{dev set}. At this time the Unlearning process would restart. In this way, the Unlearning and retraining alternate until the PPL on the target corpus $\mathcal{D}_T$, reaches the endpoint (which is described below), and the GRACE algorithm would be ended.

%This involves retraining on samples from the \emph{non-target} corpus. Following the dataset partitioning methodology of \citet{together2023redpajama1}, we construct a non-target dataset. 
In practice, the retraining set is constructed by randomly sampling instances from the rest of RedPajama dataset \cite{together2023redpajama1} after excluding the target corpus.
For instance, if we aim to unlearn the C language, the retraining set is set to be a random subset of the remainder of the RedPajama dataset after excluding the C language portion. Note that, to increase the diversity of the retraining dataset and prevent model performance degradation on unrelated domains, at each round of retraining, we would resample 30, 000 new instances. 
%each task's $non-target$ dataset is configured to contain a minimum of 30, 000 data entries. This ensures that the retraining dataset is sufficiently large, allowing each $non-target$ sample to be retrained no more than once.

\textbf{Endpoint of the Unlearning Process} 
A critical issue of the Machine Unlearning algorithm is when to stop the Unlearning process, so that the forgotten model $M_u^T$ can approximate the state as if the original model $M$ has never seen the target corpus $\mathcal{D}_T$. Prior methods achieve this by case study \cite{eldan2023s} or manually selecting certain corpus $\mathcal{D}_S$ that is highly similar to $\mathcal{D}_T$, whereas $M$ has never been trained on it. So that the performance of $M$ on the unlearned dataset $\mathcal{D}_S$ can be taken as the endpoint of Unlearning on $\mathcal{D}_T$. However, since the data filtering process of LLMs is opaque, it is hard to find a specific corpus that the model has not been trained on for each kind of target corpus. %Moreover, due to the diversity of pretraining corpora, it would be rather laborious to manually select a $\mathcal{D}_S$ for each target dataset.

To address this issue, we propose a randomized text-based method. Specifically, given an instance from $\mathcal{D} _T$, we tokenize it and
%select a text length of the first 1024 tokens from the target samples, 
randomly split the tokens into pieces with a length range from 1 to $n$, and then we shuffle their order and paste the shuffled pieces into a randomized text.  The endpoint of the Unlearning process is defined to be the point that the $M_T^u$'s PPL on $\mathcal{D} _T$ equals $M$'s PPL on the randomized text.
%Rather than seeking for a certain unseen corpus similar to $\mathcal{D} ^T$, we shuffle $\mathcal{D} ^T$ into randomized texts and define the endpoint of the Unlearning process as the point that the $M_T^u$'s PPL on $\mathcal{D} _T$ equals $M$'s PPL on the randomized text. 

This is because: (1) Through randomizing, the knowledge, semantic, and logical relationship within $\mathcal{D}_T$ are disrupted, hence, if the PPL of $M_T^u$ on $\mathcal{D}_T$ is close to the PPL of $M$ on the corresponding randomized text, it suggests that $M_T^u$ has completely forgotten $\mathcal{D}_T$; 
 (2) Compared to $\mathcal{D}_T$, the randomized text shares a similar lexical distribution, which would eliminate the influence of domain-specific vocabulary distribution.

\begin{figure} [t]
\centering
% \tiny
\setlength{\belowcaptionskip}{-10pt}
\includegraphics[width=0.8\linewidth]{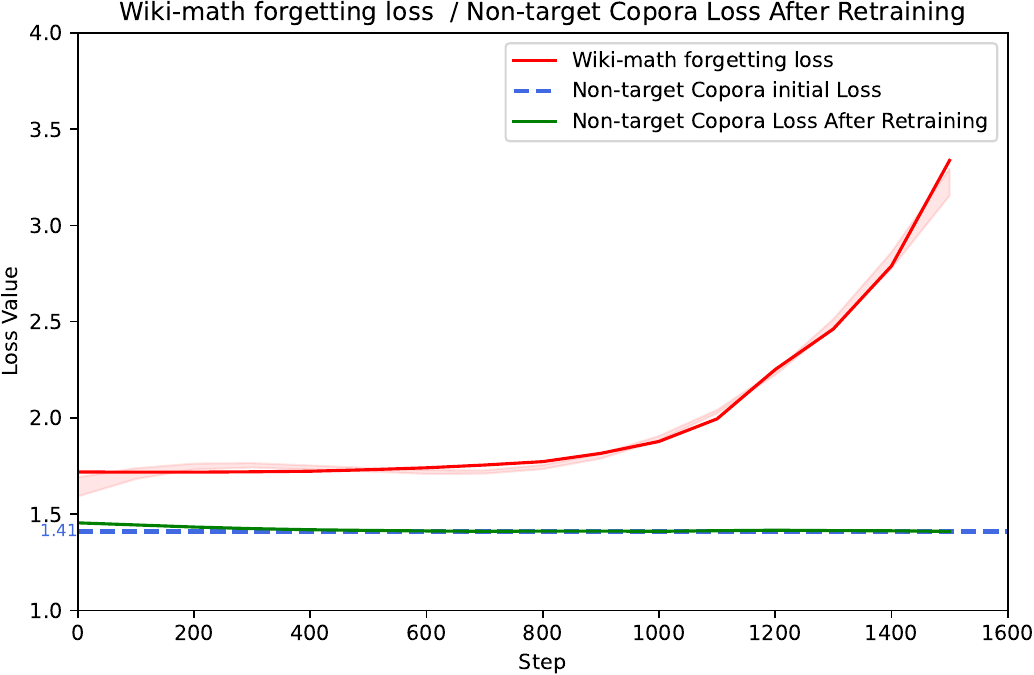} 
\caption{Comparative Analysis of Model Unlearning Effects between the $target_{math}$ and $non\text{-}target_{math}$.}
\label{Figure 3} 
\end{figure}

\begin{table} [t]
\centering
\setlength{\belowcaptionskip}{-10pt}
\begin{tabular} {>{\centering\arraybackslash} p{1cm} >
{\centering\arraybackslash} p{1.8cm} >
{\centering\arraybackslash} p{1.8cm} >{\centering\arraybackslash} p{1.8cm} } 
% \begin{tabular} {c c c c} 
\hline
Name & Mathematics & Physics &Chemistry\\ 
\hline
Before  & 5.60   & 5.48  & 5.47 \\ 
After   & 28.14   & 10.64 &7.75  \\\hline
Name & Biology & Economics &History  \\\hline
Before  & 5.66   & 5.51  & 5.41\\ 
After   & 6.97   & 6.66 &6.06  \\\hline
Name & Psychology & Law &Linguistics \\ \hline
Before  & 6.15   & 4.69  & 5.85\\ 
After   & 7.05   & 5.08 &5.97  \\\hline
\end{tabular} 
\caption{Perplexity of LLama on subsets of the Wikipedia corpus after Unlearning the Math subset of Wikipedia.} 
\label{table:Wiki_Unlearning} 
\end{table} 

\subsection{Validity Analysis of GRACE} 

Previous analyses demonstrate the effectiveness of gradient ascent-based Machine Unlearning methods in eliminating certain knowledge of LLMs \cite{eldan2023s}. We conduct further analyses to show the effectiveness of GRACE in Unlearning certain domains of knowledge and certain kinds of reasoning abilities, and the precision of not incurring unwanted impacts.

%\subsubsection{Knowledge Unlearning} 

\begin{figure*} [t]
\centering
\tiny
\setlength{\belowcaptionskip}{-10pt}
\includegraphics[width=\textwidth]{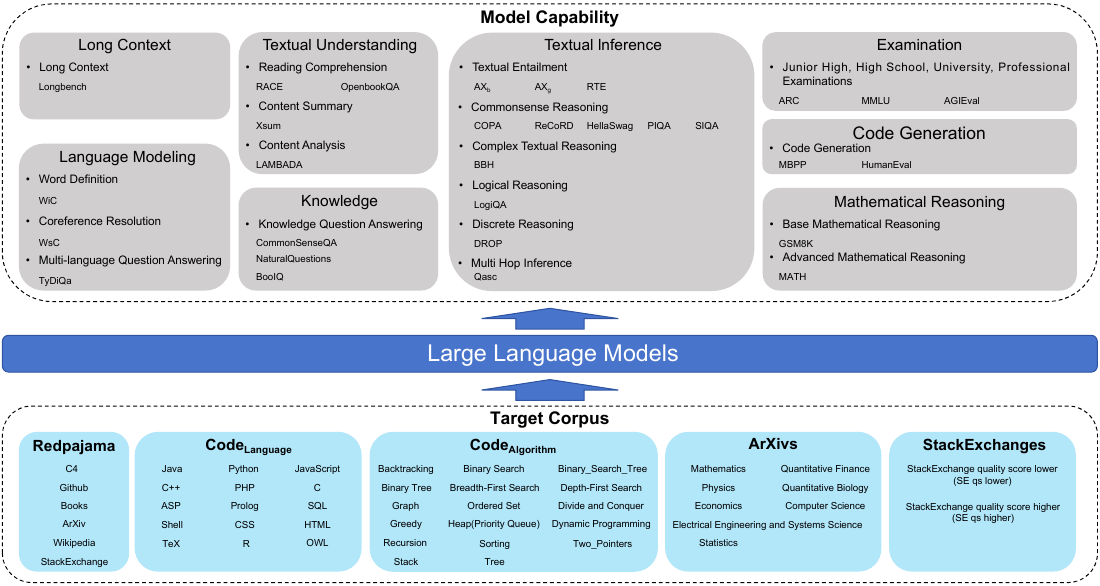}
\caption{The overall framework of the experiment.} 
\label{main} 
\end{figure*}

\textbf{Experimental Settings} 
To validate the precision and effectiveness of our methodology in selectively Unlearning certain domains of knowledge, we conduct Machine Unlearning using GRACE and take the mathematical subset of the Wikipedia corpus \cite{together2023redpajama1} as the target corpus $\mathcal{D} _{\text{math} } $. For the non-target dataset, we randomly select samples from the rest part of Redpajama corpus \cite{together2023redpajama1} which explicitly excluded  $\mathcal{D} _{\text{math} } $. 
To analyze the impact of GRACE, we not only evaluate the performance on the targeted domain $\texttt{math}$, but also include potentially related domains $\texttt{physics}$, $\texttt{chemistry}$, and $\texttt{biology}$  and unrelated domains: $\texttt{economics}$  $\texttt{history}$, $ \texttt{psychology}$, 
 $\texttt{law}$ and$ \texttt{linguistics}$.  Experiments are conducted with Llama-2-7B \cite{touvron2023llama2}, more details are provided in Appendix ~\ref{Experiment details:appendix}.

\textbf{Analyses} 
We demonstrate the loss curve on the target corpus and the non-target corpora in Figure~\ref{Figure 3}, and the final PPL of the forgotten model $M_u$ on each domain in Table~\ref{table:Wiki_Unlearning}.
%The process of model training and the corresponding experimental outcomes are illustrated in Figure 3. 
It can be observed that the loss of the model on $\mathcal{D} _{\text{math} }$ continuously increases during the Unlearning process. In contrast, due to the additional retraining process of GRACE, there is no significant increase in the loss for the non-target data. This suggests that GRACE would not incur unintentional model performance on the non-target domains.
%, thereby substantiating the precision of our approach.
%To further elucidate the model's performance variations before and after the Unlearning process, additional tests were conducted, with the results detailed in Table 1.
Moreover, as shown in Table~\ref{table:Wiki_Unlearning}, after the Unlearning process, the performances on the $\texttt{physics}$, $\texttt{chemistry}$, $\texttt{biology}$ and $\texttt{economics}$ domains, demonstrate a degradation, while the model performance upon $\texttt{history}$, $\texttt{psychology}$, $\texttt{law}$ and $\texttt{linguistics}$, remain unaffected. 
Interestingly, the extent of performance degradation in these domains is consistent with human cognition about the relevance of these domains with math: $\texttt{physics}$ and $\texttt{mathematics}$ are rather closely related; $\texttt{chemistry}$, $\texttt{biology}$, and $\texttt{economics}$ share certain common grounds with $\texttt{mathematics}$. In contrast, the correlation between historical, psychological, legal, and linguistic knowledge with mathematical knowledge is quite limited. 
%Consequently, there is no significant change in the model's performance on datasets related to these areas, before and after Unlearning. 
These observations suggest that GRACE can eliminate certain domains of knowledge from LLMs without involving unwanted impacts, indicating the effectiveness and precision of our proposed method. In the Appendix~\ref{case:appendix}, we provide more evidences about the validity of our analysis method.

% In the Unlearning process, we \text{fine-tuned} \text{Llama-2-7b} with a context length of 1024, a learning rate of $2^{\text{-} 5} $, a batch size of 4, and 40 gradient accumulation steps. %In terms of LoRA \text{fine-tuning}, we referred to the parameter settings from the \text{pre-training} section of Chinese\-LlaMA \cite{Chinese-LLaMA-Alpaca}. The rank was set to 8, lora\_alpha to 16, lora\_dropout to 0.05, and the target\_modules were q\_proj and v\_proj.

\section{Main Analysis} 
%在验证方法有效性后，我们进行实验以分析各个数据对于模型性能的影响。其中, 3.1介绍实验设置，3.2介绍各类数据如何孤立影响模型性能，3.3介绍各类数据对模型性能的联合影响。
After the validity analysis, we employ GRACE to investigate the impact of various corpora on the performance of LLMs. Specifically, Section 3.1 introduces the experimental settings. Section 3.2 explains how different types of data affect model performance individually. Section 3.3 discusses the joint impact of various types of data on the abilities of LLMs.

\subsection{Experimental Settings} 

\subsubsection{Target Corpora} 

% A critical issue is choosing and categorizing the target corpora.
%In other words, which kinds of datasets should we investigate. 
Since the ultimate goal of data influence analysis is to provide empirical guidance for optimizing the organization of the pretraining corpus of LLMs, among various open-sourced datasets, we focus our study on various subsets of the Redpajama dataset \cite{together2023redpajama1}, a replication version of the pretraining corpus of Llama \cite{touvron2023llama1,chen2023skill,fu2024hardware}. Moreover, considering the importance of complex reasoning ability, to further investigate the source of such ability, we include a set of programming algorithmics, as they can be viewed as an abstraction of thought patterns.
As shown in Table~\ref{table:Target_Data}, these datasets have been further divided into subsets, a total of 48 distinct datasets are chosen as target corpora. Specifically:

%For each target corpora, a subset with 2000 instances is randomly sampled the ``Unlearning set''. %To precisely evaluate the impact of various data types on experimental performance, this study meticulously tested 51 distinct datasets. The specific data types are detailed in  Appendix \ref{Target:appendix}.

%In selecting datasets from Github, Arxiv, Books, C4, Wiki, and StackExchanges, this paper adhered to the research methodology of Redpajama \citet{together2023redpajama1}. Given the vast scale of the Redpajama dataset, this study conducted random sampling from the aforementioned datasets, extracting 2, 000 samples from each category for a Unlearning  these target-data.

% \begin{itemize} [leftmargin=*]
% [label=\textbullet,itemsep=1pt, parsep=1pt, leftmargin=*, itemindent=0em,topsep=0pt]
\begin{itemize}[label=\textbullet,itemsep=0pt, parsep=0pt, leftmargin=*, itemindent=8pt,topsep=0pt]
    \item  All subsets of \textbf{RedPajama} \cite{together2023redpajama1}, including: C4, Github, Books, ArXiv, Wikipedia, StackExchange. These corpora play pivotal roles in the pretraining corpus of various LLMs \cite{ren2023context,zhang2024tinyllama}.  %Considering the messy content and the prevalence of low-quality instances in Web texts, temporarily in our study we do not include the CommonCrawl corpus.
      \item  The \textbf{ArXivs} contains eight subsets of the Arxiv dataset, as listed in Figure ~\ref{main} 
     \item  The \textbf{StackExchanges} dataset is obtained by dividing the StackExchange portion of the Redpajama.This subsets into two subsets based on the number of ``likes'' each Q\&A pair has received. Intuitively, the more likes an answer receives, the more likely it is to be a high-quality answer.

    \item  \textbf{$\text{Code} _\text{Algorithm} $} contains 17 kinds of important leetcode algorithm problems \cite{hartford2023leetcode}. 
      \item  The  \textbf{$\text{Code} _\text{Language} $} dataset is derived from the GitHub corpus, encompassing 15 types of programming language, spanning a variety of programming paradigms including Object-Oriented and Procedure-oriented languages, Declarative languages, Scripting languages, Front-end languages, as well as unctional language.
     % \item The \textbf{ArXiv} dataset is built based on the Arxiv part from Redpajama, containing eight categories, i.e.,. 
     % \item  The \textbf{StackExchanges} dataset is obtained by dividing the StackExchanges portion of the Redpajamathis subsets into five subsets based on the number of ``likes'' each Q\&A pair has received. Intuitively, the more likes an answer receives, the more likely it is to be a high-quality answer.
\end{itemize} 
We provide more details about the target corpora in the Appendix ~\ref{Target:appendix}.

\subsubsection{Evaluation Benchmarks} 

To comprehensively evaluate the influence of target corpora on LLMs' performance, following \citet{2023opencompass, eval-harness}, as shown in Figure \ref{main}, we select a totally of 31 benchmarks covering 9 major ability of tasks and 21 sub-categories of capabilities. Details about these datasets and the experimental settings are provided in Appendix \ref{test:appendix}.

\subsubsection{Model for Analysis} 

% We conduct all experiments with the widely adopted open-source decoder-based generative LLMs \text{Llama-2-7B} model. 
We conducted all experiments using the widely adopted open-source decoder-based generative LLMs Llama-2-7B. The reasons lie in that: (1) llama2-7B is large and powerful enough to represent LLMs; (2) The training process of the llama2-7B is typical; and (3) The existence of the Scaling Law \cite{kaplan2020scaling} allows us to infer the impact of various types of data on larger models by examining their effects on the Llama2 model.

\subsection{Impact of Individual Corpus on Model Capabilities} 

\textbf{Analysis Method} 
\label{5.1} 
As the difficulty of benchmarks is different, to make the performance changes on these benchmarks comparable, we first normalize them to the \emph{performance degradation ratio}, which is defined as \small $\gamma_{i, u} = \frac{ A_{i, j} ^{u}-A_{j} ^{o}} {A_{j} ^{o} } $\normalsize, where $A_{i, j} ^{u} $ is the model's performance on task $j$ after unlearning the $\text{target-data}_i$, and $A_{j} ^{o}$ represents the performance of the original model on task $j$. In the below, we measure the impact brought by Machine Unlearning using the \emph{performance degradation ratio}.

\begin{figure*} [t]
\centering
\setlength{\belowcaptionskip}{-8pt}
\includegraphics[width=\textwidth]{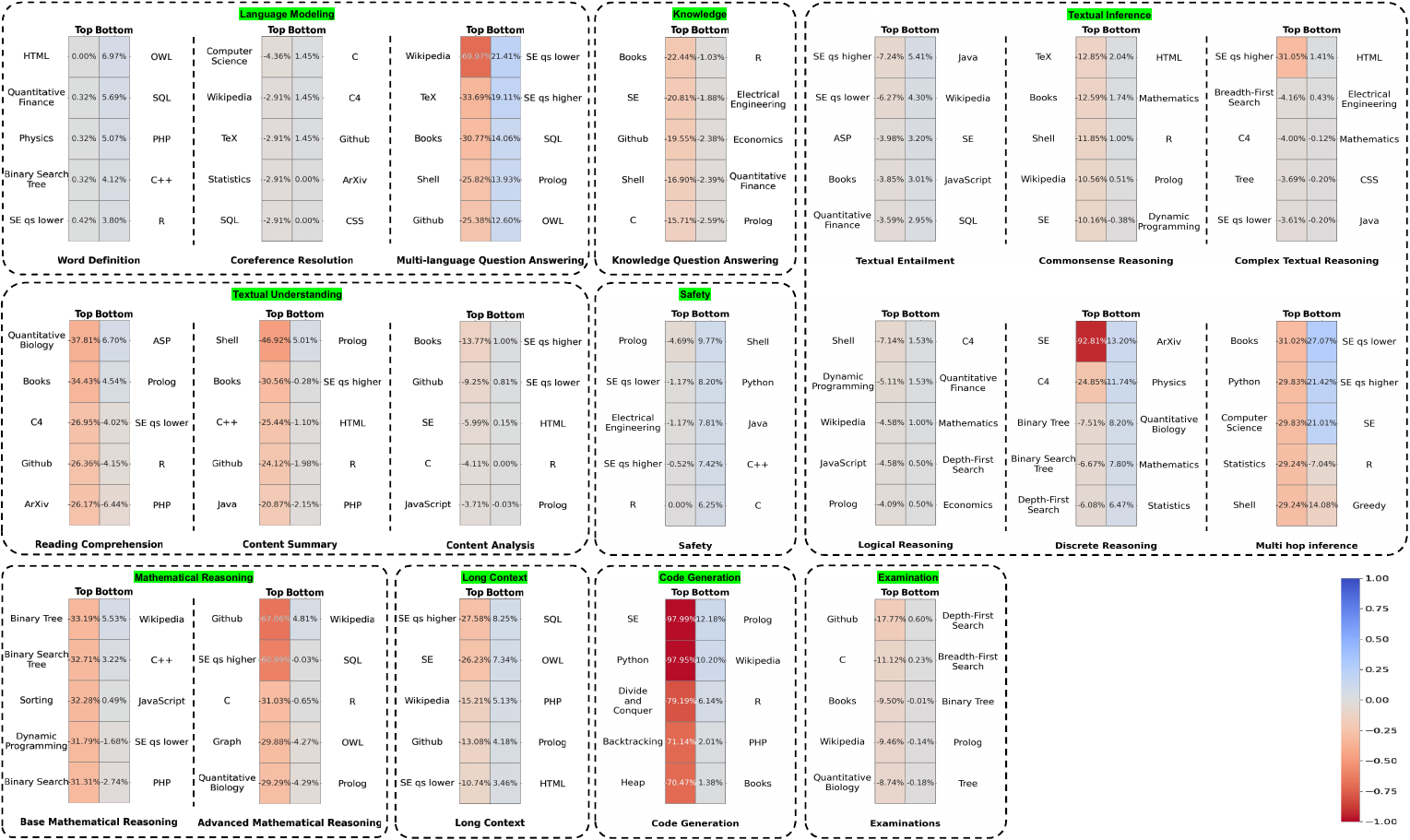} 
\caption{The Top and Bottom 5 datasets that have the most and the least impact on each type of model capability.} 
\label{result} 
\end{figure*} 

\textbf{Analysis Results} 
Figure~\ref{result} lists the Top and Bottom 5 datasets that have the most and the least impact on each type of model capability. 
%The term ``average'' refers to the model's average performance after Unlearning 51 datasets. It is important to note that the target data currently selected exhibit negative gains in the ``Word Definition'' capability. This suggests that the ability to ``Word Definition'' may potentially stem from a broader range of datasets
From which we can observe that:

% \textbf{Language Modeling} 
% 算法对语言建模能力的影响超过代码语言本身
% \begin{itemize} [leftmargin=*, align=left]
% \begin{itemize} [leftmargin=*, align=left]
\begin{itemize} 
 [label=\textbullet,itemsep=0pt, parsep=0pt, leftmargin=*, itemindent=8pt,topsep=0pt, partopsep=0pt]
\item 
\textbf{Language Modeling} As a fundamental of LLMs, the language modeling ability is seldom significantly impacted by a specific type of corpus alone. As an (approximate) multi-lingual parallel corpus, the Wikipedia corpus may play a critical role in aligning different languages for an LLM, and influence the multilingual ability of LLMs.
% The logical aspect of algorithms exerts a more pronounced influence on the model's language modeling capabilities than the code language itself. Given the disparity between the distribution of code and natural text, the model's language modeling ability may even see an improvement after a significant portion of the code is forgotten.
% The logical aspect of algorithms has a more significant impact on the model's language modeling capability than the language itself.It is noteworthy that the distribution of code differs significantly from natural text; hence, the language modeling capability of the model may even improve after forgetting a majority of the code.

% Code, as a complete language system, enables the model to acquire multilingual abilities through systemic transfer learning. This explains why the Llama \cite{touvron2023llama1} model can process Russian to some extent, despite the absence of Russian in its training corpus. Unlearning the coding component disrupts the linguistic structure of the model, leading to a loss of multilingual capabilities.

\item \textbf{Textual Understanding} 
One prominent phenomenon is that programming language corpora have a high impact on the textual understanding ability of LLMs. Heuristically, codes are abstractions of relationships between real-world objections and could be helpful for understanding the semantic and logical relationships among text. Moreover, knowledge-rich corpora such as books and Arxiv, and corpora with high diversity such as books and C4 profoundly influence the textual understanding ability of LLMs. 
%There exists a discernible relationship between the diversity of text and comprehension, aligning with the findings presented in \cite{longpre2023pretrainer}. For example, datasets such as the Books corpus, which encompass a wide array of information types, exert a more substantial influence on the model's comprehension abilities compared to datasets comprised of homogeneous text types.
Hence, corpora with diversity, rich commonsense knowledge and code corpus may constitute three foundations for the textual understanding ability of LLMs.

\item \textbf{Textual Inference} 
Text inference tasks depend on a wide range of corpora. Notably, besides commonsense-related corpora, text reasoning tasks also extensively rely on various types of code corpora, algorithm corpora, and mathematical corpora. For example, tasks like Big Bench Hard \cite{suzgun2022challenging} significantly depend on high-quality StackExchange content and algorithms such as breadth-first search. This demonstrates the importance of symbolic reasoning capabilities represented by mathematics and code in understanding the deep logical relationships within texts.

% Highly ``likes''StackExchange content is crucial for various reasoning tasks, particularly Challenging Reasoning. The coding sections significantly contribute to most textual reasoning tasks. 
% such as dynamic programming and binary tree play critical role in , respectively. 
% Due to the distinct distributions between natural and coding languages, code has a negative effect on Textual Entailment tasks. The BOOK dataset also positively influences the model on tasks requiring common sense reasoning, which is in accordance with \cite{longpre2023pretrainer, shen2023slimpajamadc}. Conversely, \text{target-datasets} like Wikipedia and ArXiv have limited or even negative impacts on various reasoning tasks.

\item \textbf{Knowledge Reasoning} Datasets such as  Books, and  are invaluable for solving real-world knowledge problems. Programming languages, as an integral part of the global knowledge system, significantly influence the model's knowledge capabilities. The ArXivs dataset, due to its complexity and deviation from common world knowledge, has a lesser impact. 
% Algorithms, as a form of abstract knowledge, have limited influence on concrete world knowledge but their degradation could be attributed to the Unlearning of linguistic capabilities.

\item \textbf{Mathematical Reasoning} 
The source of LLMs' math reasoning capabilities has drawn great attention from researchers, as it could be an indicator of the complex reasoning ability of LLMs \cite{ernest2023abduction}. From the results in Figure~\ref{result}, high-quality mathematical texts and code corpora (especially algorithms) have a significant impact on LLMs' math reasoning abilities. This demonstrates: (1) There is a close relationship between mathematical and coding abilities \cite{soldaini2024dolma, shen2023slimpajamadc}. To some extent, both math and code problems are abstractions of real-world problems, and involve complex symbolic reasoning processes to solve them. Hence, model performances demonstrate high sensitivity upon the algorithmic. (2) High-quality mathematical texts are key to the model learning of mathematical abilities.

\item \textbf{Code Generation} 
Compared to text-based tasks, the range of knowledge for code generation tasks is relatively narrow, only limited to mathematics and code-related corpora \citet{shen2023slimpajama}. This again demonstrates the close relationship between mathematics and coding. Overall, forgetting algorithmic knowledge has a greater impact on the model's coding ability than specific programming languages. This suggests that the model's understanding of algorithms does not depend on specific program languages. In other words, LLMs could understand the logic of algorithms, instead of memory algorithmic knowledge depending on certain programming languages.

\item \textbf{Long Text} GitHub, Wikipedia, Books, and  ``High-liked'' StackExchanges significantly impact the model's long text capabilities, as these corpora are composed of long texts, entailing complex logical relationships and abundant common sense knowledge.

%Shorter texts like algorithms have limited impact due to their brevity.

\item \textbf{Examination} Completing exam questions requires extensive knowledge and strong reasoning abilities. Hence, code, commonsense, mathematics, and books corpora all form the foundation of an LLMs's examination capabilities.

\item \textbf{Safety} 
Interestingly, the model's security is enhanced after forgetting the code corpora. This might be due to the absence of emotional factors in the code corpora and its straightforward logic, thus making the generated results more crude and aggressive.

\end{itemize} 
% \end{itemize} 
More detailed experimental data and results can be found in Appendix ~\ref{Experimental Result:appendix}.

\subsubsection{Corpus with Broad Influences} 

We calculate corpora that can influence multiple capabilities. Since they may lead to a broad influence on model capabilities, these datasets may serve as the foundation of the training corpus.

% \textbf{Analysis Method} In the target-data, certain datasets significantly influence numerous capabilities of the model. We adopted the same method for data processing as described in Section ~\ref{5.1}. Furthermore, we define a dataset as "heatmap data" if, upon its Unlearning, the decline in performance across 70\% of the capabilities exceeds the average decline observed across all datasets.

% Within the target data, certain datasets significantly impact many of the model's capabilities. We adopted the same method of data processing as in Section ~\ref{5.1}, while also defining datasets that influence over 70\% of the capability types beyond the average value as ``heatmap data''.

% \textbf{Analysis Results} 

% \text{target-data}, four categories—Books, Shell, C and StackExchange quality score 4 meet the criteria for heatmap data. Unlearning Books and Shell datasets results in a decline in 15 capabilities beyond the average, whereas Unlearning C and StackExchange quality score 4 leads to a decline in 14 capabilities beyond the average. These datasets significantly impact multiple model capabilities and play a crucial role in model training.
 
% Among the selected target data, the primary ``heatmap data''datasets include Books, C, StackExchange quality score 4, and Shell. The Books and Shell datasets impact 15 types of capabilities, whereas C and StackExchange quality score 4 influence 14 types. These datasets have a substantial effect on multiple capabilities of the model and play a crucial role in its training process.
\textbf{Analysis Method} In the \text{target-data}, certain datasets significantly influence numerous capabilities of the model. We adopted the same method for data processing as described in Section ~\ref{5.1}. Furthermore, we define a dataset as ``High-impact data'' if its removal leads to a performance decline in over 70\% of capabilities, which exceeds the average decline observed across all datasets.

\textbf{Analysis Results} Among the selected \text{target-data}, four categories—Books, Shell, and Github meet the criteria for High-impact data. Unlearning Books datasets result in a decline in 16 capabilities beyond the average, whereas Unlearning Shell and GitHub leads to a decline in 14 capabilities beyond the average. These datasets significantly impact multiple model capabilities and play a crucial role in model training.

\subsection{Joint Impact of Multiple Corpora on Model Capabilities} 

Previous research indicates that LLMs can combine information from multiple kinds of corpora, and generalize it to new tasks. Hence, during the pretraining stage, multiple corpora may perform joint impact upon LLMs. In this section, we explore measuring the joint contribution of multiple datasets upon the LLMs.

% 好多领域, 遗忘之后性能上升的, 也和咱们说能力之间有冲突相匹配 注意！！！

\begin{figure} [t]
\centering
\setlength{\belowcaptionskip}{-10pt}
\includegraphics[width=\linewidth]{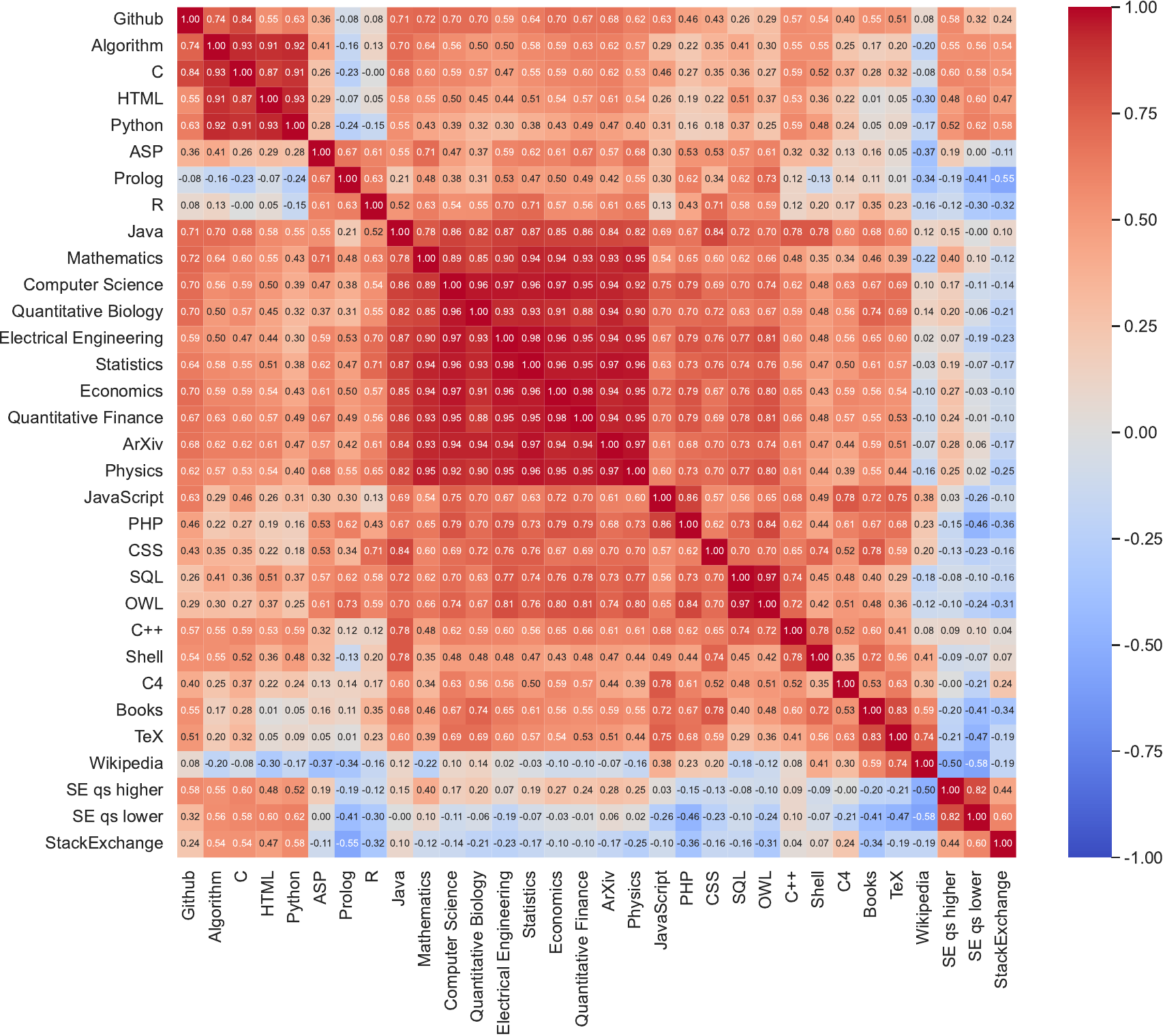} 
\caption{A correlation matrix based on the model's performance across 19 capabilities after experiencing data Unlearning. Among them, ``Algorithm'' is the average value of all Algorithm.} 
\label{Interrelationships} 
\end{figure} 

\subsubsection{Interrelationships Among Data} 

\textbf{Analysis Method} 
To explore the relationships between datasets from different domains, we calculated the Pearson correlation coefficient between any two target corpus $T_A$ and $T_B$, based on the model's performance degradation ratio upon 19 capabilities after Unlearning $T_A$ and $T_B$. Within each category of abilities, each subtype was given an equal weight. When a subtype included multiple datasets, we averaged the performance changes across these datasets to calculate the coefficients. These coefficients form a correlation matrix. If the correlation coefficient between two datasets is positive, it indicates that the impact of these two datasets on downstream tasks is similar, and vice versa. Subsequently, we conducted hierarchical clustering on the correlation matrix to categorize datasets. Figure ~\ref{Interrelationships} displays the correlation matrix rearranged according to the categories defined by the hierarchical clustering.

\textbf{Analysis Results} 
As shown in Figure~\ref{Interrelationships}, according to their relationships, the corpora could be categorized into three types, which we name as ``Correlated Corpora'', ``Complementary Corpora'', and ``Orthogonal Corpora'', respectively. Specifically:
% \begin{itemize} [leftmargin=*, align=left]
\begin{itemize}[label=\textbullet, itemsep=0pt, parsep=0pt, leftmargin=*, itemindent=8pt, topsep=0pt]
    \item   \textbf{Correlated Corpora} refers to corpora that the model has similar performance changes after Unlearning them. In other words, they have a similar influence on LLMs. For instance, the correlation coefficients among Economics, Quantitative Finance, and Statistics are all greater than 0.95. Hence, to some extent, the correlated corpora can substitute for each other in the training corpus, leading to redundancy in the training corpus and a waste of computation resources. Hence, it would be necessary to reorganize these corpora to enhance pre-training efficiency.

    % It is noteworthy that previous studies by \citet{} have posited that GitHub and StackExchanges exhibit similar performances in mathematical reasoning and code generation. However, our findings indicate discernible differences between the two. Specifically, GitHub exerts a more significant influence on the model's mathematical reasoning capabilities, whereas StackExchanges have a greater impact on the model's code generation abilities. This distinction underscores the nuanced roles different data sources play in enhancing specific aspects of model performance in the context of natural language processing and computational tasks.
    \item   \textbf{Complementary Corpora} refers to corpora that the model performance alternations are different after Unlearning them.
    Interestingly, our analyses suggest that there may exist certain corpora that have a complementary influence on the model's performance. 
    For example, the  Wikipedia corpus and SE-qs-lower corpus have a negative correlation coefficient of -0.58. As detailed illustrated in Appendix ~\ref{Experimental Result:appendix},
    this negative correlation is because both Wikipedia and SE qs lower have impacts on multiple capabilities, while the capabilities influenced by these two datasets seldom overlap. Thus, these two corpora could act simultaneously as a critical composition of the pretraining corpus. In general, the math corpora (e.g., StackExchange) have a complementary relationship with commonsense-related corpora such as Books or Wikipedia.  
    Note that our results also suggest the existence of an extreme case that the involvement of one dataset can cause a decline in the performance of another dataset. This situation requires further verification. If it indeed exists, then in organizing the pre-training data, a trade-off must be made between the two datasets.
    \item\textbf{Orthogonal Corpora} refers to two corpora having a correlation coefficient near zero. For example, the correlation coefficient between Wikipedia and ArXiv is -0.07. This suggests that these corpora independently contribute to the model's different capabilities with low redundancy. Hence, when optimizing the organization of the pretraining corpus, each of the orthogonal corpora should not be recklessly excluded to avoid impairing the comprehensiveness of the pre-training dataset.

\end{itemize} 

\textbf{Discussions} 

\textbf{(1) About scaling law:} Such complementary and conflicting relationship also indicates that the term ``scaling law'' refers to the \textbf{scaling law under the same data composition}. This is because, \textbf{the coefficient of scaling law would be influenced by the composition of the pretraining corpus}. For example, on Correlated Corpora, the model loss may decrease fast due to the synergistic effect of datasets. Hence, \emph{a lower loss level or faster loss decrease does NOT necessarily indicate the superiority of a training algorithm or a model architecture}.  

\textbf{(2) About groups of the pretraining data:} Moreover, according to the correlation matrix, there seem to be several groups of data: (1) Math-related group composed of StackExchange-related corpora; (2) Knowledge-related group including the subsets of Arxivs (e.g., Statistics / Economic). Note that, group (1) and (2) seems to be complementary; (3) The intermediate group between (1) and (2) composed of C, python, prolog, etc., bridges (1) and (2). More groups could be identified by further dividing the pretraining corpus.

\begin{figure}[ht]
\centering
\includegraphics[width=0.99\linewidth]{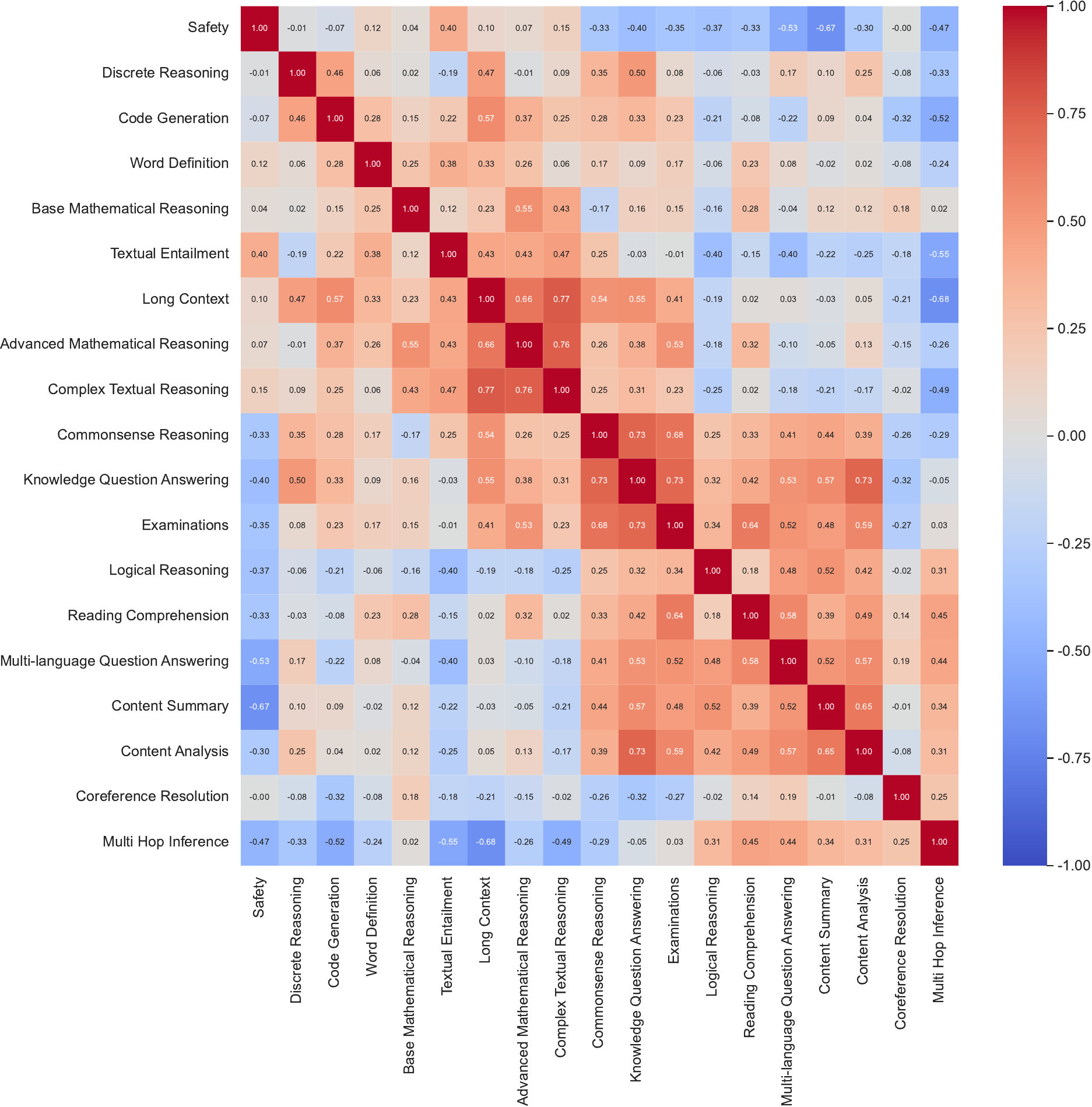} 
\caption{A correlation matrix based on the model's performance across 48 Unlearning tasks after experiencing data Unlearning.} 
\label{ability} 
\end{figure} 

\subsubsection{Interrelationships Among Model Capbilities} 

Additionally, for an arbitrary kind of capability benchmark, we can obtain a model performance after forgetting each category of pretraining data. Hence, using a similar methodology, we can calculate a correlation matrix between the capabilities of an LLM and analyze the relationship patterns between model abilities. Owing to space constraints, the results are presented in Figure \ref{ability}. From which we can (clearly) observe three groups of abilities: (1) \textbf{Textual modeling-related abilities}, such as Coreference Resolution, Content summary, and Reading Comprehension; (2) \textbf{(Symbolic) Reasoning-related abilities}, such as Base Mathematical Reasoning, Code Generation, and Long Context Modeling. (\emph{Interestingly, long context modeling shows a strong correlation with code generation and math reasoning. This indicates that the essential difficulty in long context modeling may be under the complex relationship among the context}); (3) \textbf{Composition of the above two kinds of abilities}, such as Advanced Mathematical Reasoning, Complex Textual Reasoning, and Examinations. These results suggest that there are categories of fundamental abilities of LLM and the complex or advanced abilities are essentially built upon these categories of fundamental abilities.

Moreover, it could be observed that these abilities can be both positively correlated (synergistic) and negatively related (antagonistic). For example, the Advanced Mathematical Reasoning ability is positively related to code generation and Long Text Modeling, indicating that enhancing these two kinds of abilities will also enhance the Advanced Mathematical Reasoning ability of an LLM, and vice versa, demonstrating a synergistic relationship between these three kinds of abilities. However, multi-hop knowledge inference shows a negative correlation relationship between several reasoning-related tasks, such as code generation, and mathematical reasoning. This indicates that there are inherent conflicts between certain capabilities. For example, \emph{increasing the commonsense-related ability would be at the cost of reasoning ability}. 

The relationship patterns of the contribution of different datasets and the relationship patterns model abilities highlight the necessity of optimizing the pretraining corpus's organization to enhance the pretraining process's efficiency and effectiveness.

\section{Hints about Optimization of the Pretraining Corpus and the Training Process}

The above results imply several suggestions for optimizing the organization of the pretraining corpus and training process:

\subsection{About the Proportion of Each Category of Pretraining Corpus} To some extent, \emph{if the change in an LLM's capability is more sensitive to a certain category of data, it indicates that this category of data is more important for enhancing this kind of capability, and therefore, its proportion should be increased accordingly}. In Figure 4 and Table 8-10 of this paper, we demonstrate the sensitivity of model capability on the ablation of each category of the pretraining corpus in detail. For example, as Figure 4 shows, to increase the Advanced Mathematical Reasoning ability of an LLM, the GitHub corpus, Stack Exchange Corpus, and certain algorithms such as Graph Algorithm may be of vital importance and should be involved more, while Wikipedia, SQL, R, OWL, and Prolog have minimal influence upon the Advanced Mathematical Reasoning ability of an LLM, and should not be overinclusive. Moreover, in this sense, \textbf{the ``high-energy'' data, such as books, GitHub, etc., should be included in the training corpus as much as possible}.

This also highlights \textbf{the necessity of subdividing the pretraining datasets, as they may indeed bring different capabilities for the LLM and should have different weights rather than be treated as a whole mixture}. For example, further subdividing the GitHub corpus according to the type of programming language. This is because different programming languages may correspond to different programming paradigms and different modes of thinking, have different application scenarios, and would naturally bring different types of capabilities to the model.

\subsection{About the Interrelationship between Corpora} As Figure 5 shows, there are relationship patterns between different corpora, such as the correlated relationship. Such relationship patterns should also be considered when deciding the proportion of each kind of category of corpus. For example, for highly positively correlated corpora such as the Mathematics part of Arxiv and Computer Science and Arxiv, their weights in the pretraining corpus should be decreased because they can largely substitute each other. Incorporating the correlation patterns into the consideration would increase the information density of the pretraining corpus, and thus increase the efficiency and effectiveness of the pretraining process.

Moreover, the correlation matrix of pretraining corpora can be decomposed into several orthogonal ``principle components'', with these components explaining most of the variation of the correlation matrix, and hence these principal components can be explained as a sort of ``pivotal data'' composed by the weighted mixture of several different components. Construct the pretraining corpora according to the principal components would  

\subsection{About the Arrangement of Dataset During the Pretraining Process} In the classical LLM training paradigm, all instances within the pretraining corpus randomly appear in the pretraining process. However, as Figure 5 and Figure 6 indicate, there are dependency relationships between the abilities of LLM, for example, the Advanced Mathematical Reasoning ability depends on the Base Mathematical Reasoning ability. It would be impractical to learn a subsequent complex capability without having sufficient prerequisite abilities and knowledge. Hence, \textbf{it might be necessary to introduce time order for training data in the pretraining process}, such as arranging instances corresponding to complex abilities in the later stage of the whole pretraining process. For example, place the corpus about long-context modeling or advanced mathematical ability in the later stage of pretraining. In other words, \textbf{different datasets should appear in different stages of the pretraining process}.

Recently, \citet{minicpm2024} introduced a Two-stage training paradigm by arranging data with more complex tasks into the second stage of pretraining and observed additional performance improvements. This further indicates the reasonability of our conclusions and calls for further investigations about how should the pretraining data be ordered in the time sequential. 
 
\subsection{About the Evaluation of Pretraining Process} Since the loss level and the speed of loss decrease would be influenced by the composition of the dataset, it would be necessary to eliminate the influence of data composition, such as separately evaluating the loss curve on each category of data, rather than assessing the loss value on the whole development set that mixes all types.

We plan to investigate the validity of these suggestions through de novo training an LLM in the following works.

\section{RELATED WORK} 

%\subsection{Data Influence Analysis} 
The Data Influence Analysis (DIA) task aims at finding how each training data contributes to a model's performance.
DIA methods can be mainly classified into two categories \cite{hammoudeh2022training}.
% The first category, Retraining-Based approaches, involves assessing influence by training models with and without specific instances. 
The first category, the Retraining-Based approach \cite{jia2021scalability, kandpal2022deduplicating, ghorbani2019data}, assesses the influence of certain instances by comparing the model performance with and without these instances. 
% This category encompasses methods like Leave-One-Out \cite{jia2021scalability}, DownSampling \cite{kandpal2022deduplicating}, and Shapley Value techniques \cite{ghorbani2019data}.
However, due to the prohibitive training costs, these methods have only been extensively applied in the `small'' models 
% \cite{jia2021scalability,  nguyen2023bayesian}.
\cite{jia2021scalability,ghorbani2019data,kandpal2022deduplicating,nguyen2023bayesian}.

% The subsequent category, Gradient-Based methods, focuses on comparing gradient similarities between training and test instances.
The second category Gradient-Based methods \cite{koh2017understanding, koh2019accuracy, pruthi2020estimating, hara2019data} discover training samples with greater influence by comparing gradient similarities between training and test instances.
% and includes techniques such as Static Gradient-Based Influence Estimation \cite{koh2017understanding, koh2019accuracy} and Dynamic Gradient-Based Influence Estimation \cite{pruthi2020estimating, hara2019data}. 
These methods demonstrate efficacy in finding instances contributing to knowledge memorization of models. However, they may fail to find instances related to the reasoning abilities of models, which is especially important for LLMs, as it may originate from groups of correlated instances that jointly contribute to the performance of LLMs. 
% For example, solving math problems requires understanding a knowledge taxonomy, and the taxonomy is described by a set of interdependent instances holistically. Missing one component would lead to the collapse of the whole taxonomy. Hence, the gradient-based methods may fail to trace the influence of such a whole corpus, which is of vital importance \cite{grosse2023studying}. 
Considering the limitations of these methods, we propose a Machine Unlearning-based DIA approach to investigate the impact of corpora upon LLMs.

Machine Unlearning is devised to erase certain knowledge from a model. 
Prior research \cite{jang2022knowledge,graves2021amnesiac,gupta2021adaptive,sekhari2021remember} suggests that machine unlearning can selectively erase specific knowledge from a model through gradient \emph{ascent} on corresponding instances.
 \citet{eldan2023s} has shown that gradient ascent methods would still be effective in LLMs, and can accurately unlearn targeted samples. In this paper, we further extend the gradient ascent-based machine unlearning methods by involving an additional retraining process and a random text-based stop criterion.

\section{Conclusion} 
In this study, we employed a Machine-Unlearning-based data influence analysis method GRACE to investigate the complex effects of diverse types of pretraining data on the performance of Large LLMs. We gained empirical analysis results about how specific components of the pretraining corpus influence LLMs capabilities, and how they jointly contribute to multiple capabilities of LLMs. Our findings suggest the nuanced impact of data selection and organization in LLMs development. The identification of high-impact data and the delineation of complementary, antagonistic, and orthogonal data relationships offer guidance for optimizing pre-training data organization.
In future work, we consider adapting our analysis methodology to other parts of LLMs training such as supervised fine-tuning.

\section{Limitations}

This study systematically investigated the impact of various pre-training datasets on the capabilities of LLMs using the GRACE method. While our findings offer valuable insights into the subtle relationship between pre-training data and model capabilities, it is crucial to acknowledge the inherent limitations of our approach.

\subsection{Data Limitations}
There remains a still remains space for exploration in the domain of data. In terms of breadth, our study primarily focused on the Redpajama dataset and its subsets. However, there are other datasets for LLMs, and Redpajama may not include all the corpora encountered during the training process of LLMs. In terms of depth, some important data domains may still possess ample subdivision space. For example, Books datasets can be segmented by books type, which could limit the comprehensiveness of our analysis.

% \subsection{Model Limitations}
% Due to computational resource constraints, our research concentrated on the Llama-2-7B model. The extent to which our findings are applicable to other LLMs or models of different scales remains an open question. Factors such as the model's architecture, training objectives, and the scale of pre-training data might influence the impact of pre-training datasets.

\subsection{Limitations of Evaluation Metrics}
The existing evaluation systems might not adequately unearth the deeper capabilities of models, potentially overlooking subtle variations and thereby missing valuable insights.

\subsection{Future Research Directions}

Addressing these limitations provides opportunities for future research. Using a broader variety of data and applying our analytical framework to other models are key steps toward a more comprehensive understanding of the relationship between pre-training data and LLMs capabilities. These efforts will contribute to the ongoing discussions about optimizing pre-training strategies to enhance model performance and efficiency.

%\bibliography{anthology, custom} 
\bibliography{custom} 
\clearpage
\onecolumn
\appendix
\section{Experiment details} 
\label{Experiment details:appendix} 
All experiments were conducted on an A100-SXM4-80GB cluster, with all models utilizing the bf16 precision format.For each target corpora, a subset with 2000 instances is randomly sampled the ``Unlearning set''
\section{Example Appendix} 
\label{case:appendix}

% \subsubsection{Interrelationships Among Data} 
\begin{figure} [t]
\centering
\includegraphics[width=\linewidth]{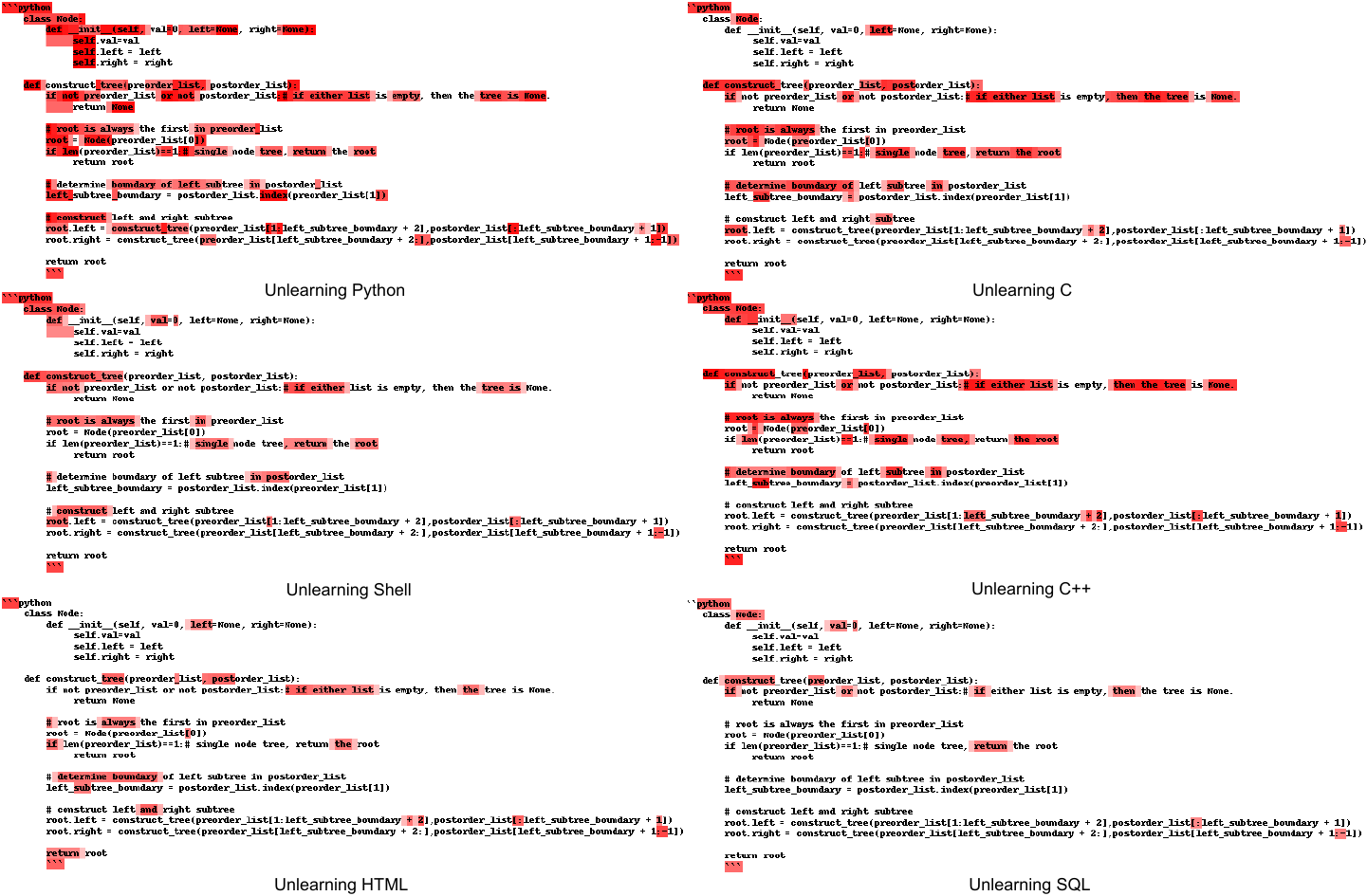} 
\caption{The performance of the model on an unaltered Python statement after selectively Unlearning programming languages, including Python, C, Shell, C++, HTML, and SQL.} 
\label{case1} 
\end{figure} 

\textbf{Analysis Method} 
As illustrated in Figure ~\ref{case1}, we evaluate the performance of the model on an unaltered Python statement after selectively Unlearning programming languages, including Python, C, Shell, C++, HTML, and SQL. The darker the color, the greater the increase in token loss compared to the original model. 

\textbf{Analysis Results} Notably, the most significant loss increase is observed after the model unlearns Python, indicating a profound impact on its ability to interpret Python statements. Furthermore, when Unlearning other languages, the changes in model loss predominantly occur in comments and certain keywords, which are syntactically or semantically shared across these languages. Languages like C++, C, and Shell, which structurally resemble Python to some extent, exhibit substantial loss variations. Conversely, HTML and SQL, being more distinct from Python, result in minimal changes in the model's performance on Python statements after being forgotten. Additionally, we observe that due to the object-oriented features of Shell and C++, Unlearning these languages leads to a notable increase in the loss associated with the ``class'' token. In contrast, given C's procedural nature, the loss related to ``class'' does not show significant variation. This analysis underscores the intertwined nature of programming language knowledge within the model and highlights the differential impact of Unlearning specific languages on the model's comprehension of Python code.

\begin{figure} [t]
\centering
\includegraphics[width=\linewidth]{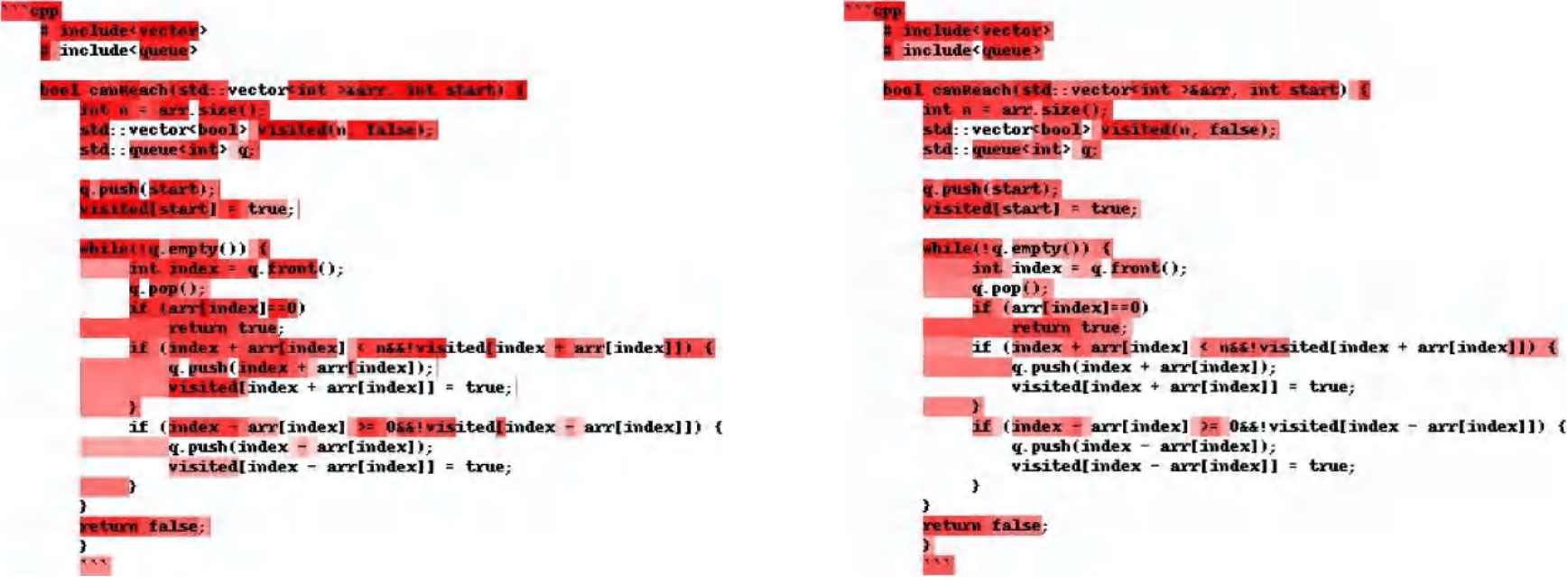} 
\caption{The performance of the model on an unaltered BFS statement after selectively Unlearning programming languages, including DFS and Graph.} 
\label{case2} 
\end{figure} 
\textbf{Analysis Method} 
% As illustrated in Figure ~\ref{case2}, we evaluate the performance of the model on an unaltered Python statement after selectively Unlearning programming languages, including Python, C, Shell, C++, HTML, and SQL. The darker the color, the greater the increase in token loss compared to the original model. 
As illustrated in Figure ~\ref{case2}, we evaluate the model's performance on Breadth-First search statements written in C++ after selectively applying Unlearning algorithms, including depth-first and graph algorithms. The darker the color, the greater the increase in token loss compared to the original model
\textbf{Analysis Results} As illustrated in Figure ~\ref{case2}, the performance of the model post-omission of the Depth-First Search (DFS) algorithm on Breath First Search (BFS) data is depicted at the right, while the performance post-omission of the Graph algorithm on BFS data is shown at the left. It is observable that there is a significant increase in loss for both cases in terms of variable definitions and certain key terms. However, during the execution phase of the BFS algorithm, the loss in the model after omitting the Graph algorithm is substantially greater than that after omitting the DFS algorithm.
\section{Test DaseSet} 
\label{test:appendix} 

In Table 2, we integrate the frameworks \cite{2023opencompass, eval-harness} to construct a comprehensive evaluation system for large models.

Language Modeling. The average performance on WiC \cite{pilehvar2019wic}, WSC \cite{levesque2012winograd}, and TyDiQa \cite{clark2020tydi} is reported. These test sets are all evaluated with 0-shot results.

Knowledge. The average performance is reported using BooIQ \cite{clark2019boolq}, CommonSenseQA \cite{talmor2018commonsenseqa}, and NaturalQuestions \cite{kwiatkowski2019natural}. We report 8-shot results for CommonSenseQA and 0-shot results for all other benchmarks.

Textual Inference. The average performance is reported using $\text{AX} _{b} $ \cite{wang2020superglue}, $\text{AX} _{g} $ \cite{wang2020superglue}, RTE \cite{wang2020superglue}, COPA \cite{roemmele2011choice}, ReCoRD \cite{zhang2018record}, HellaSwag \cite{zellers2019hellaswag}, PIQA \cite{bisk2019piqa}, SIQA \cite{sap2019socialiqa}, BBH \cite{suzgun2022challenging}, LogiQA \cite{liu2020logiqa}, DROP \cite{Dua2019DROPAR}, and Qasc \cite{khot2020qasc}. We report 3-shot results for BBH, 2-shot results for DROP, and 0-shot results for all other benchmarks. Notably, $\text{AX} _{b} $, $\text{AX} _{g} $, and RTE are used for Textual Entailment tasks; COPA, ReCoRD, HellaSwag, PIQA, and SIQA for Commonsense Reasoning tasks; BBH for Complex Textual Reasoning; LogiQA for Logical Reasoning; DROP for Discrete Reasoning; and Qasc for Multi-hop Inference.

Mathematical Reasoning. The average performance is reported using GSM8K \cite{cobbe2021gsm8k} and MATH \cite{hendrycksmath2021}. GSM8K represents Base Mathematical Reasoning, while MATH represents Advanced Mathematical Reasoning. These test sets are all evaluated with 4-shot results.

Textual Understanding. The average performance is reported using RACE (Middle and High) \cite{lai2017race}, OpenbookQA \cite{OpenBookQA2018}, Xsum \cite{narayan2018dont}, and LAMBADA \cite{paperno2016lambada}. RACE (Middle and high) and OpenbookQA are used for Reading Comprehension tasks, Xsum for Content Summary tasks, and LAMBADA for Content Analysis tasks. These test sets are all evaluated with 0-shot results.

Long Context. The model's ability in long text understanding and reasoning is represented using English datasets within Longbench \cite{bai2023longbench}.

Code Generation. We report the average pass@1 scores of models on HumanEval \cite{chen2021codex} and MBPP \cite{austin2021program}. MBPP is evaluated with 1-shot results, while HumanEval is evaluated with 0-shot results.

Examination. The average performance is reported using ARC (easy and challenge) \cite{clark2018think}, MMLU \cite{hendryckstest2021, hendrycks2021ethics}, and AGIEval \cite{zhong2023agieval}. We report 5-shot results for MMLU and 0-shot results for all other benchmarks. It is worth noting that for AGIEval, we only selected English datasets.

Safety. Performance is represented using TruthfulQA, evaluated with 0-shot results.
\begin{table} [t]
\small
\centering
\begin{tabular} {>{\centering\arraybackslash} p{4cm} >{\centering\arraybackslash} p{5.5cm} >{\centering\arraybackslash} p{4.5cm} } 
\hline
\multicolumn{2} { c } {Ability} & Test DaseSet
 \\ \hline
\multirow{3} {*} {Language Modeling} & Word Definition & WiC (0\text{-} shot) \\
&Coreference Resolution & WSC (0\text{-} shot) \\
&Multi-language Question Answering & TyDiQa (0\text{-} shot) \\
\hline
\multirow{3} {*} {Knowledge } &
\multirow{3} {*} {Knowledge Question Answering} & BooIQ (0\text{-} shot) \\
&& CommonSenseQA (8\text{-} shot) \\
&& NaturalQuestions (0\text{-} shot)\\
\hline

\multirow{12} {*} {Textual lnference} & \multirow{3} {*} {Textual Entailmen} & $\text{AX} _b$ (0\text{-} shot)\\
&& $\text{AX} _g$ (0\text{-} shot)\\
&& RTE (0\text{-} shot) \\ 
\cline{2-3} 
&\multirow{5} {*} {
Commonsense Reasoning
} & COPA (0\text{-} shot)\\
&& ReCoRD (0\text{-} shot)\\
&& HellaSwag (0\text{-} shot) \\ 
&& PIQA (0\text{-} shot) \\ 
&& SIQA (0\text{-} shot) \\ 
\cline{2-3} 
&Complex Textual Reasoning  &BBH (3\text{-} shot)\\
&Logical Reasoning &LogiQA (0\text{-} shot)\\
&Discrete Reasoning 
&DROP (2\text{-} shot)\\
&Multi hop inference
&Qasc (0\text{-} shot)\\
\hline
\multirow{2} {*} {Mathematical Reasoning
 } &
Base Mathematical Reasoning & GSM8K (4\text{-} shot) \\
&Advanced Mathematical Reasoning
& MATH (4\text{-} shot) \\
\hline
\multirow{4} {*} {Textual Understanding
 } &
\multirow{2} {*} {Reading Comprehension
} & RACE (0\text{-} shot) \\
&& OpenbookQA (0\text{-} shot) \\
\cline{2-3} 
&Content Summary &Xsum (0\text{-} shot)\\
&Content Analysis
&LAMBADA (0\text{-} shot)\\
\hline
\multirow{2} {*} {Long Context} & Long Context Understanding & \multirow{2} {*} {Longbench} \\
& Long Context Reasoning &  \\
\hline
\multirow{2} {*} {Code Generation} &
\multirow{2} {*} {Code Generation} & MBPP (1\text{-} shot) \\
&& HumanEval (0\text{-} shot) \\
\hline
\multirow{3} {*} {Examination} &
\multirow{3} {*} {\parbox{5.5cm} {\centering Junior High, High School, University, Professional Examinations} } 
& ARC (0\text{-} shot) \\
&& MMLU (5\text{-} shot) \\&& AGIEval (0\text{-} shot) \\
\hline
Safety &Safety &TruthfulQA (0\text{-} shot)\\
\hline

\end{tabular} 
\caption{Classification of the Test Data.} 
\label{table:Test_Data} 
\end{table} 
\section{Target DaseSet} 
\label{Target:appendix} 
The \textbf{$\text{Code} _\text{Algorithm} $} corpus contains building on the leetcode dataset created by \citet{hartford2023leetcode}, selected 17 of the most important algorithms. Following the methodology of \citet{selfinstruct}, for algorithm types with not enough size, we utilized GPT-4 \citet{openai2023gpt4} for data augmentation to ensure a minimum of 2, 000 samples per algorithm. Furthermore, to ensure the model's Unlearning pertains to the algorithms themselves and not to a specific programming language, each algorithm problem was represented in five language formats: C++, Python, Java, JavaScript, and pseudocode.

% The  \textbf{$\text{Code} _\text{Language} $ } dataset is derived from the GitHub portion of the Redpajama dataset \citet{together2023redpajama1}.This dataset undergoes a stringent filtration process, selectively including data where a single programming language constitutes over 99.99\% of the content and the aggregate length surpasses 2000 bytes. This process is designed to isolate target data pertinent to each specific language. Subsequently, a random assemblage of 2000 samples per language is aggregated to form the comprehensive $\text{Code} _\text{Language} $ dataset.

The \textbf{$\text{Code} _\text{Language} $} dataset is derived from the GitHub portion of the Redpajama dataset \citet{together2023redpajama1}. 
In terms of language selection, based on the characteristics of programming languages, this study chose 15 types, including Object-Oriented, Procedure-oriented, Declarative programming, Scripting, frontend and other common languages.
It is specifically filtered to include data where a single language proportion for more than 99.99\% of the content and the total length is in excess of 2000 bytes, aiming to isolate target data for a specific language. 
From this filtered data, 2000 samples are randomly selected for each language to constitute the $\text{Code} _\text{Language} $ dataset.

The \textbf{Arxivs} dataset is constructed based on the Arxiv part of the Redpajama dataset \citet{together2023redpajama1}, which is further divided into eight main categories. Randomly select 2000 samples from each category to form the Arxiv dataset.

The \textbf{StackExchanges} dataset is constructed from the StackExchanges segment of the Redpajama dataset \citet{together2023redpajama1}, comprising Q\&A pairs that have garnered more than 5 ``likes''. Subsequently, it is segregated into two tiers based on the count of ``likes''. The underlying rationale is that the number of ``likes'' is indicative of an answer's quality and popularity. Consequently, these five subsets encompass samples ranging in quality and popularity from the lowest to the highest.

\begin{table} [t]
\centering
\begin{tabular} {>{\centering\arraybackslash} p{2.5cm} |>{\centering\arraybackslash} p{4cm} >{\centering\arraybackslash} p{4cm} >{\centering\arraybackslash} p{4cm} } 
\hline

Target Data & \multicolumn{3} { c } {name} \\ \hline
\multirow{2} {*} {Redpajama} 
  & C4 &Github &Books \\  
  &ArXiv &Wikipedia &StackExchange
     \\ 
\hline
\multirow{5} {*} {$\text{Code} _\text{Language} $} 
&C &C++ &CSS\\ &HTML &Java &Python \\&PHP &JavaScript &Shell\\ &R &Web Ontology Language&SQL \\&TeX &ASP &Prolog\\ 
\hline
\multirow{6} {*} {$\text{Code} _\text{Algorithm
} $} 
&Backtracking &Binary Search	&Binary Search Tree	\\&Binary Tree	&Breadth\-First Search	&Depth\-First Search \\	
&Divide and Conquer	&Dynamic Programming	&Graph	\\&Greedy	&Heap (Priority Queue)	&Ordered Set	
\\&Recursion	&Sorting	&Stack	\\&Tree	&Two Pointers\\
\hline
\multirow{3} {*} {ArXiv} 
&Physics	&Mathematics	&Computer Science	\\&Quantitative Biology &Quantitative Finance &Statistics \\&Electrical Engineering\\
\hline
\multirow{2} {*} {StackExchanges}  &StackExchange quality score lower &StackExchange quality score higer \\
\hline
\end{tabular} 
\caption{Classification of the \text{Target-Data.} } 
\label{table:Target_Data} 
\end{table} 

\section{Experimental Result and Joint Impact of Multiple Ability on Model Capabilities} 
\label{Experimental Result:appendix} 
\begin{figure} [ht]
\centering
\includegraphics[width=\linewidth]{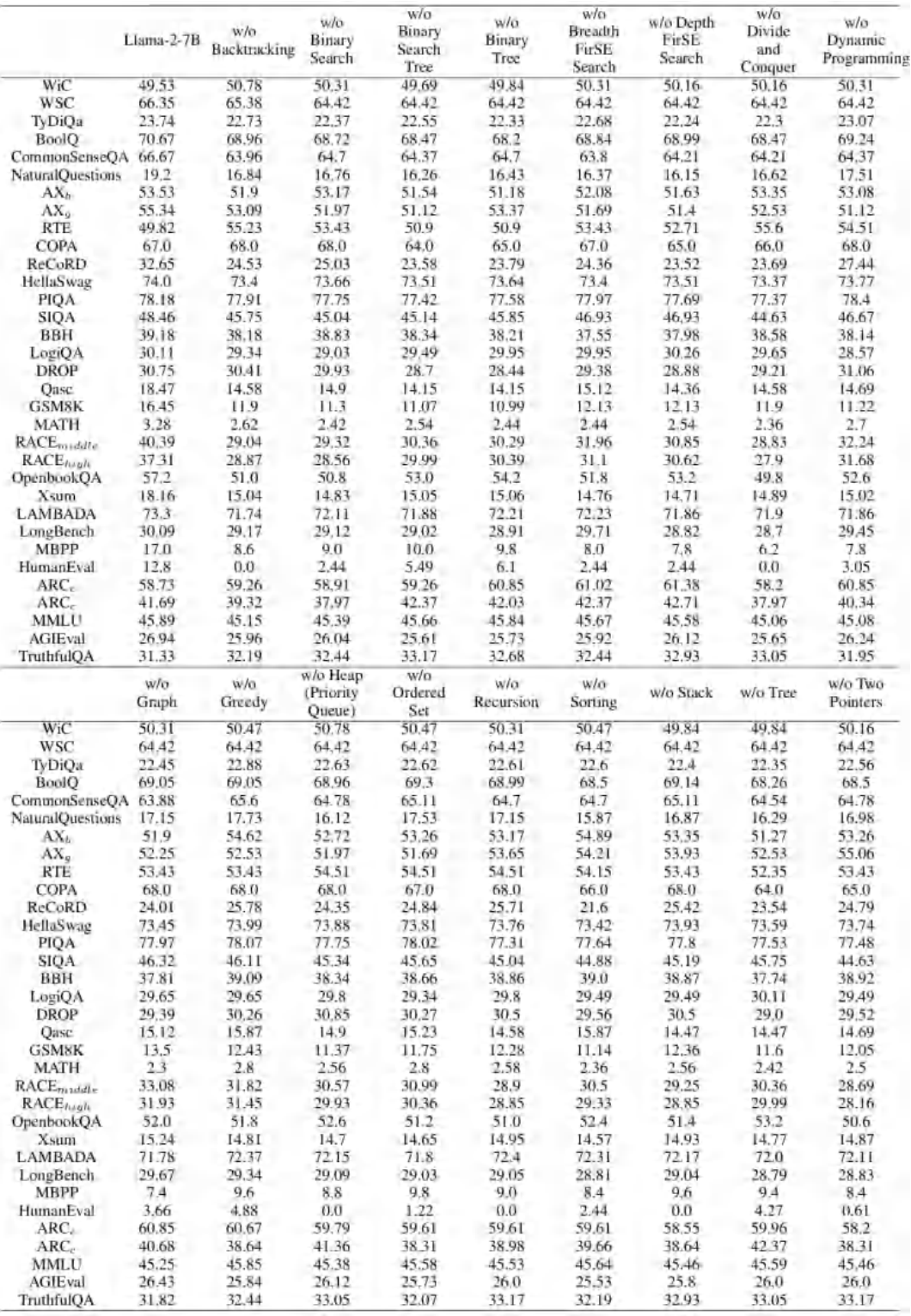} 
\caption{The first part of results for Unlearning different types of datasets} 
\end{figure} 

\begin{figure} [ht]
\centering
\includegraphics[width=\linewidth]{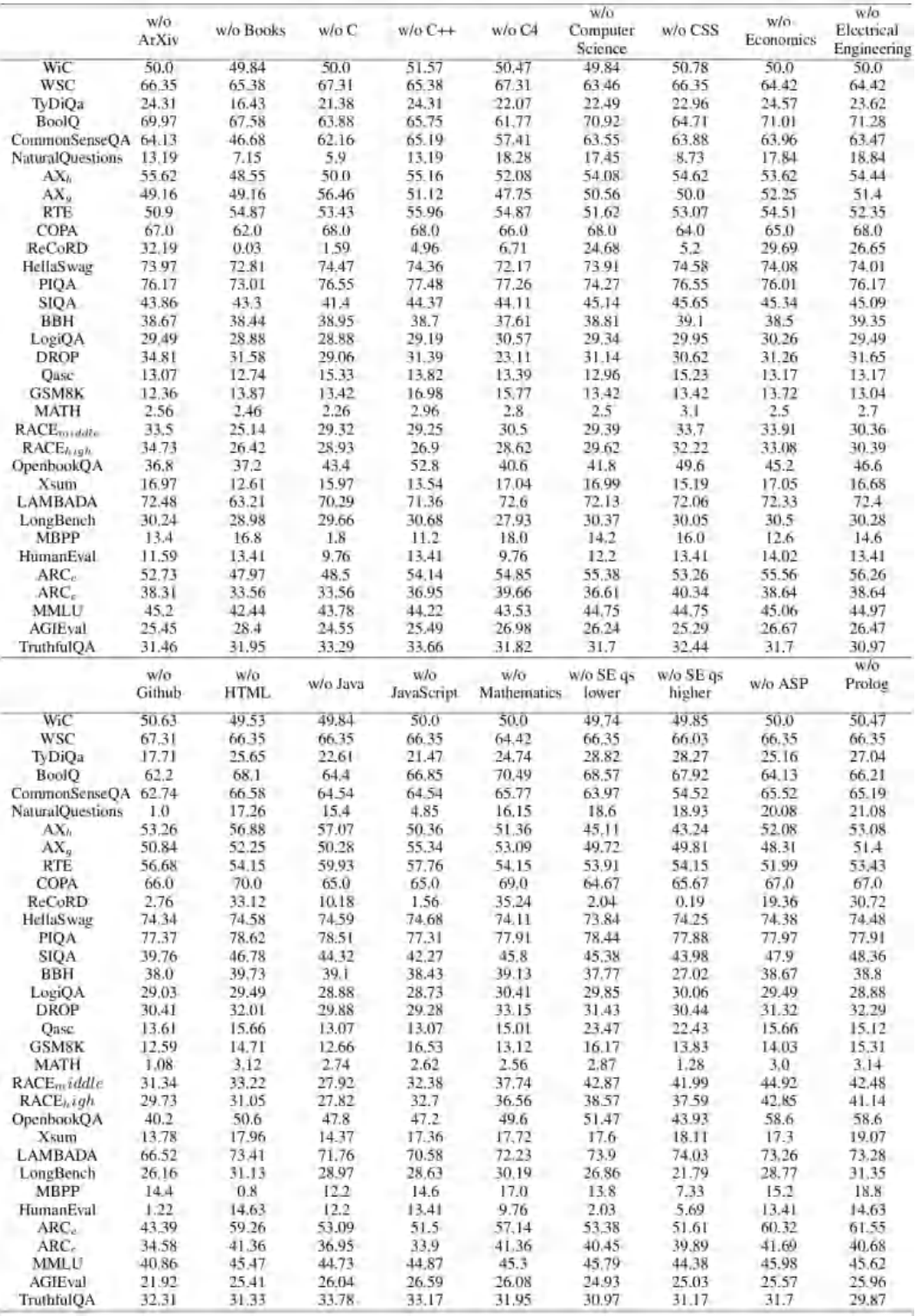} 
\caption{The second part of results for Unlearning different types of datasets} 
\end{figure}

\begin{figure} [ht]
\centering
\includegraphics[width=\linewidth]{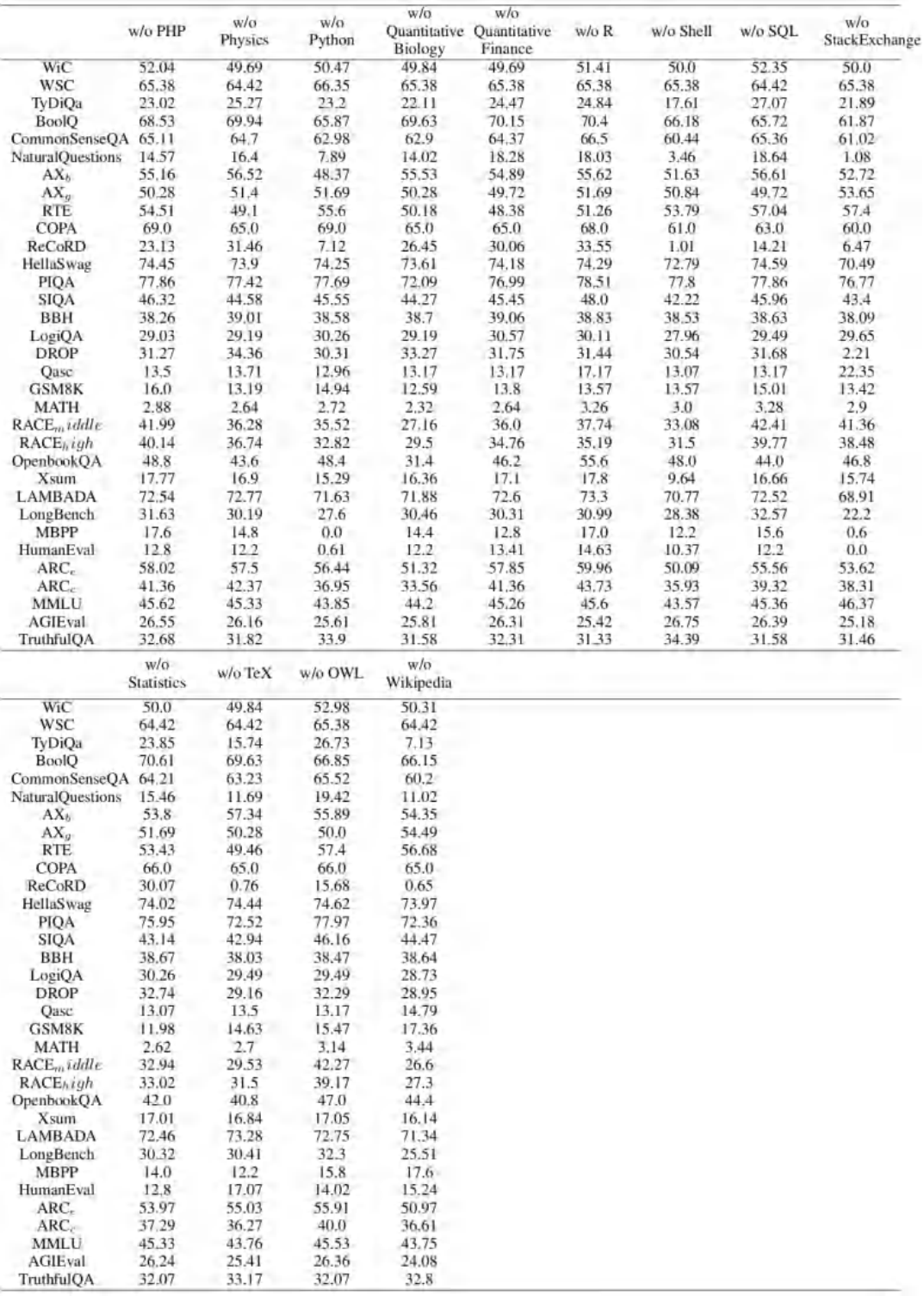} 
\caption{The third part of results for Unlearning different types of datasets} 
\end{figure}

\end{document}